%% file: main.tex
\documentclass[10pt,twocolumn,letterpaper]{article}

\usepackage{cvpr}              
\usepackage{algorithm}
\usepackage[noend]{algpseudocode}
\usepackage{algorithmicx}
\usepackage{amsmath}
\usepackage{subcaption}
\usepackage{bbding}
\usepackage[accsupp]{axessibility}  

\definecolor{cvprblue}{rgb}{0.21,0.49,0.74}
\usepackage[pagebackref,breaklinks,colorlinks,allcolors=cvprblue]{hyperref}

\def\paperID{946} 
\def\confName{CVPR}
\def\confYear{2025}

\newcommand{\mcal}{\mathcal}

\usepackage{mdframed}

\newcommand{\be}{\begin{equation}}
\newcommand{\ee}{\end{equation}}
\definecolor{Gray}{gray}{0.85}
\definecolor{LightCyan}{rgb}{0.88,1,1}

\newcommand{\bfx}{\mathbf{x}}

\newcommand{\bfz}{\mathbf{z}}
\newcommand{\bfI}{\mathbf{I}}

\newcommand{\bfzero}{\mathbf{0}}

\definecolor{blue1}{RGB}{0,128,255}
\definecolor{blue3}{RGB}{0,0,128}
\definecolor{darkpastelgreen}{rgb}{0.01, 0.75, 0.24}
\definecolor{cerulean}{rgb}{0.0, 0.48, 0.65}

\definecolor{darkgreen}{rgb}{0,0.6,0}

\title{Z-Magic: \underline{Z}ero-shot \underline{M}ultiple \underline{A}ttributes \underline{G}uided \underline{I}mage \underline{C}reator}

\author{Yingying Deng$^{\circ}$,Xiangyu He$^{\circ,1}$,Fan Tang$^{\diamondsuit,2}$, Weiming Dong$^{1}$\\
$^{1}$ MAIS, Institute of Automation, Chinese Academy of Sciences \\
$^{2}$ Institute of Computing Technology, Chinese Academy of Sciences\\
{\tt\small dyy15@outlook.com, tfan.108@gmail.com, weiming.dong@ia.ac.cn}
}

\begin{document}
\maketitle

\renewcommand{\thefootnote}{}
\footnotetext{
\textsuperscript{$\circ$}These authors contributed equally to this work.} 
\footnotetext{\textsuperscript{$\diamondsuit$}Corresponding author: Fan Tang.}
\input{sec/0_abstract}    
\input{sec/1_intro}

\input{sec/related_work}
\input{sec/3_method}

\input{sec/4_exp}
{
    \small
    \bibliographystyle{ieeenat_fullname}
    \bibliography{main}
}


\end{document}

%% file: sec/0_abstract.tex
\begin{abstract}
The customization of multiple attributes has gained popularity with the rising demand for personalized content creation. 
Despite promising empirical results, the contextual coherence between different attributes has been largely overlooked. 
In this paper, we argue that subsequent attributes should follow the multivariable conditional distribution introduced by former attribute creation. 
In light of this, we reformulate multi-attribute creation from a conditional probability theory perspective and tackle the challenging zero-shot setting. 
By explicitly modeling the dependencies between attributes, we further enhance the coherence of generated images across diverse attribute combinations. 
Furthermore, we identify connections between multi-attribute customization and multi-task learning, effectively addressing the high computing cost encountered in multi-attribute synthesis. 
Extensive experiments demonstrate that Z-Magic outperforms existing models in zero-shot image generation, with broad implications for AI-driven design and creative applications.
\end{abstract}

%% file: sec/1_intro.tex
\section{Introduction}
\label{sec:intro}


\begin{figure}[t]
    \centering
    \includegraphics[width=\linewidth]{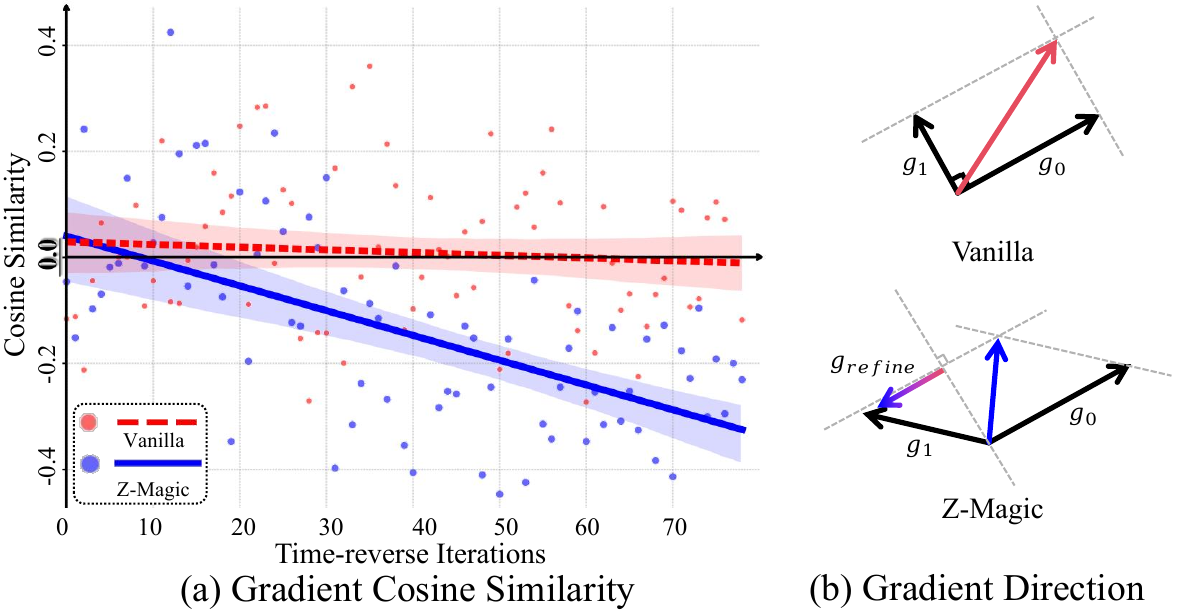}
    \caption{We visualize the cosine similarity between gradients introduced by two conditions (i.e., landmark $g_0$ and face ID $g_1$) during the diffusion process of face synthesis. In the vanilla setting, each condition is applied separately without considering contextual coherence, causing the gradient directions to be nearly orthogonal, as random vectors in high-dimensional space are often orthogonal. In contrast, our method adjusts each condition based on the preceding ones, resulting in an obtuse-angled optimization direction where later conditions refine the gradients of earlier ones. In other words, our approach has two degrees of freedom to construct the gradient direction: angle and length.}
    \label{fig:grad}
\end{figure}

With the rapid advancement of generative AI, particularly in the widespread adoption of image generation applications, the paradigm of controlled image creation has gained considerable attention from both industry and the image synthesis community~\cite{RombachBLEO22, PodellELBDMPR24, RameshPGGVRCS21, abs-2204-06125}. In this context, guiding the image creation process through multimodal conditions, such as text prompts and reference images, becomes more popular. 

In reality, conditional dependence is a fundamental property in multi-attribute guided creation. For example, in fashion design, the selection of a color scheme is often influenced by the style or cut of the clothing, and vice versa; similarly, in face synthesis, the presence of gender implies facial features, and this contextual information should guide subsequent attribute choices. These examples intuitively illustrate that attributes do not exist in isolation but are mutually informative and conditionally dependent, forming a naturally coherent whole. While existing diffusion-based approaches have achieved impressive results in conditional generation~\cite{ZhangRA23, KimODKS24, ChoiKJGY21, HuangC0023, BansalCSSGGG24, 0007DRGM24, AvrahamiHGGTPLF23, HertzVFC24}, the underlying structure that inherently links multiple attributes into a cohesive creation has been rarely discussed. Besides, for the sake of practical simplicity, previous methods often assume independence among different attributes, which limits their capacity to model inter-attribute relationships. 

In the score-based conditional diffusion model, conditioning is facilitated through a gradient-based approach, formulated as  $g_{\mathbf{c}} = \nabla_{\mathbf{x}_t} \log p(\mathbf{c} | \mathbf{x}_t) $, where the gradient  $g_{\mathbf{c}}$  represents the conditioning mechanism. When each attribute is generated under the assumption of independence, the gradients associated with distinct attributes exhibit near-zero cosine similarity. As shown in Figure \ref{fig:grad}(a), for any two conditional gradients,  $g_{\mathbf{c}_0}$  and  $g_{\mathbf{c}_1}$ , during the diffusion process, we observe that  $\mathbb{E}\left[\frac{g_{\mathbf{c}_0}^T g_{\mathbf{c}_1}}{\|g_{\mathbf{c}_0}\| \|g_{\mathbf{c}_1}\|}\right] \approx 0$. This behavior aligns with the properties of random high-dimensional vectors, which are typically nearly orthogonal in high-dimensional spaces \footnote{In high-dimensional spaces, two random vectors are almost always nearly orthogonal \cite{Shewchuk1998}.}. This result is somewhat surprising, as these conditions apply to the same image, and ideally, they should exhibit a degree of contextual correlation to support coherent and harmonious synthesis.

In light of this, we present a novel approach that explicitly models conditional dependencies, ensuring each attribute is generated in the context of previously selected ones. Specifically, our method employs a conditional distribution framework, where each subsequent attribute is derived from a multivariable conditional distribution informed by the attributes generated earlier, aligning with the joint distribution  $p(\mathbf{c}_0, \mathbf{c}_1 | \mathbf{x}_t)$. Each conditionally dependent gradient progressively refines image generation. Initially, we apply  $g_0 = \nabla{\mathbf{x}_t}\log p(\mathbf{c}_0 | \mathbf{x}_t)$  to capture the influence of the first attribute, followed by  $g_1 = \nabla{\mathbf{x}_t}\log p(\mathbf{c}_1 | \mathbf{c}_0, \mathbf{x}_t) $, which adjusts the generation to align with the contextual influence of the first attribute. As shown in Figure~\ref{fig:grad}(b), this sequential conditioning introduces a gradient refinement mechanism whereby  $g_1$  partially aligns with  $g_0$  and corrects the vector length along the  $g_0$  direction, enhancing attribute coherence.

Further, while models can be fine-tuned for specific attribute combinations, training-based approach is resource-intensive and lacks scalability, particularly in the face of high-dimensional attribute spaces. 
Thus, exploring methods for zero-shot multi-attribute synthesis has the potential to unlock new avenues for practical AI-driven creative tools. 
In response to these challenges, we introduce Zero-shot Multiple Attributes Guided Image Creator (Z-Magic), an approach designed to address the complex requirements of multi-attribute customization. 
By reformulating multi-attribute synthesis as a problem of conditional dependency across attributes, Z-Magic leverages principles from conditional probability and multi-task learning to generate coherent images without extensive attribute-specific tuning. In our experiments,
this method not only reduces human labeling overhead for attribute combinations but also paves the way for harmonious customization across diverse domains.

%% file: sec/related_work.tex
\section{Related Works}
\subsection{Conditional Diffusion Model}
Recent advances in Denoising Diffusion Probabilistic Models (DDPMs) \cite{DDPM} have demonstrated exceptional capabilities in various image synthesis tasks, including image manipulation and conditional generation. Classifier-Guidance \cite{DhariwalN21} introduces an additional classifier to guide the generation toward specific categories, while classifier-free guidance models \cite{abs-2207-12598} combine conditional and unconditional models to enhance generation quality.

Several landmark architectures have further shaped this field. GLIDE \cite{NicholDRSMMSC22} employs a pre-trained CLIP model \cite{RadfordKHRGASAM21} for text-guided image synthesis, and Stable Diffusion \cite{RombachBLEO22} improves computational efficiency by performing text-conditioned denoising in latent space. ControlNet \cite{ZhangRA23} incorporates a parallel U-Net architecture to enable diverse visual conditions, including landmarks, edge maps, and skeletal structures. Similarly, T2I-Adapter \cite{MouWXW0QS24} proposes a lightweight adapter network to integrate various visual control signals.

Despite their impressive performance, these approaches demand substantial computational resources for training. In contrast, our work explores a training-free paradigm for controlling diffusion models, offering greater flexibility and broader applicability.

\subsection{Multiple Attributes Customization}
Multi-attribute control in image generation—encompassing aspects such as style, structure, textual descriptions, and semantic segmentation—has become a critical research focus in computer vision, particularly in tasks like style transfer and facial image synthesis. StyleGAN \cite{KarrasLA21} pioneered this field by introducing a style-based generator architecture that enables precise control over visual features at multiple scales, setting new standards in high-quality face image generation.

Subsequent advances have further extended the capabilities of attribute-controlled generation. HairCLIP \cite{Wei0Z0TY0Y22} built on the StyleGAN framework by integrating both textual and reference image conditioning. TediGAN \cite{XiaYXW21} advanced semantic control by combining GAN inversion techniques with style-based generation. In diffusion models, recent research has shown significant progress: for instance, \cite{LiuM0HFL0C24} achieved control over both identity and expression attributes, while \cite{KimODKS24} demonstrated the complementary strengths of diffusion models and GANs for text and visually guided face generation. Additionally, \cite{HuangC0023} introduced a multi-modal face control framework using pre-trained diffusion models. Various style transfer techniques \cite{DengHTD24, HertzVFC24, deng:2020:arbitrary,deng:2021:arbitrary,Deng:2022:stytr} have also been developed to jointly address style and content conditions.

While these methods deliver impressive results, they often face limitations in scalability for additional control conditions. To address this, we propose a unified paradigm that enables the seamless integration of multiple attribute control signals. Our framework provides a more flexible and generalizable approach to multi-attribute-controlled image generation.

%% file: sec/3_method.tex
\label{sec:method}

\section{Approach}

\subsection{Preliminaries}
Score-based diffusion models \cite{Hyvarinen05, SongGSE19, 0011SKKEP21, SongDME21}, as a unifying framework, consolidate previous methodologies in both score-based generative modeling and diffusion probabilistic modeling. These models describe the forward diffusion process $\{\mathbf{x}(t)\}_{t=0}^T$ as a stochastic differential equation (SDE). Given this forward SDE, the reverse process can be formulated by solving the corresponding reverse-time SDE in the context of controlled generation \cite{Anderson1982ReversetimeDE, 0011SKKEP21}:
\begin{align}
\label{eq_sde}
    d\mathbf{x} =\Big\{ \mathbf{f}(\mathbf{x}, t) - g(t)^2 &\big[ \nabla_{\mathbf{x}} \log p_t(\mathbf{x}) + \\
    &\nabla_{\mathbf{x}} \log p_t(\mathbf{c} | \mathbf{x}) \big] \Big\} dt + g(t) d\mathbf{\bar{w}}, \nonumber
\end{align}
where $\mathbf{f}(\mathbf{x}, t)$ denotes the drift coefficient, and $\mathbf{\bar{w}}$ represents the standard Wiener process for reverse-time dynamics. Additionally, \cite{0011SKKEP21} proposes a time-dependent score-based model, $\mathbf{s}_{\boldsymbol{\theta}^*}$, to approximate $\nabla_{\mathbf{x}_{t}}\log p(\mathbf{x}_{t})$, thereby enabling the construction and numerical solution of the reverse-time SDE. Our subsequent discussion focuses on the discretization of these stochastic dynamics.

\subsection{Multiple Attributes Guided Creation}
There are numerous numerical methods for solving SDEs mentioned above, in this section, we use the ancestral sampling adopt in the popular DDPM \cite{DDPM} to ease the derivation \footnote{Please note that our approach is not limited to DDPMs and is also compatible with other solvers.}. The sampling method serves as the solver to the reverse-time VP SDE \cite{0011SKKEP21}, formulated as
\begin{align*}
\mathbf{x}_{t-1,\mathbf{c}}&=(2-\sqrt{1-\beta_t})\mathbf{x}_t+\beta_t\nabla_{\mathbf{x}_{t}}\log p(\mathbf{x}_{t}|\mathbf{c})+\sqrt{\beta_t}\boldsymbol{\epsilon}.
\end{align*}
Given the approximation that $\nabla_{\mathbf{x}_{t}}\log p(\mathbf{x}_{t})\approx \mathbf{s}_{\boldsymbol{\theta}^*}(\mathbf{x}_t,t)$, we have 
\begin{align}
\nabla_{\mathbf{x}_{t}}\log p(\mathbf{x}_{t}|\mathbf{c})\approx\mathbf{s}_{\boldsymbol{\theta}^*}(\mathbf{x}_t,t)+\underbrace{\nabla_{\mathbf{x}_{t}}\log p(\mathbf{c}|\mathbf{x}_{t})}_{\text{controllable term}}.
\end{align}
where $s_{\theta^*}$ is the pre-trained unconditional score estimator model,  $\mathbf{c}$ refers to the set of conditions, such as text prompts, landmark points, and reference images. By expanding the single condition guidance $p(\mathbf{x}_{t}|\mathbf{c})$ to the multiple attributes guidance
\begin{align}
p(\mathbf{x}_t|\mathbf{c}_1,...,\mathbf{c}_n)=\frac{p(\mathbf{c}_1,...,\mathbf{c}_n|\mathbf{x}_t)p(\mathbf{x}_t)}{p(\mathbf{c}_1,...,\mathbf{c}_n)},
\end{align}
it contributes to more delicate manipulations on $\mathbf{x}_{t-1}$ with $\nabla_{\mathbf{x}_{t}}\log p(\mathbf{c}_1,...,\mathbf{c}_n|\mathbf{x}_{t})$.

\subsection{Conditional Independent Creation}
When facing multiple attributes prior $\mathbf{c}_i$, suppose we were considering the position of the target object or the style of the image, these conditions themselves are independent, if they are measuring the target image among different dimensions. For instance, we have a cat in the middle of the image would not tell you anything about Van Gogh's art style. Formally,
\begin{align}
p(\mathbf{c}_1,...,\mathbf{c}_n)=p(\mathbf{c}_1)p(\mathbf{c}_2)...p(\mathbf{c}_n)=\prod_{i=1}^np(\mathbf{c}_i),
\end{align}
Following the above derivation, it seems quite straightforward to assume the conditional independence given the noisy data $\mathbf{x}_t$
\begin{align*}
p(\mathbf{c}_1,...,\mathbf{c}_n|\mathbf{x}_t)=p(\mathbf{c}_1|\mathbf{x}_t)p(\mathbf{c}_2|\mathbf{x}_t)...p(\mathbf{c}_n|\mathbf{x}_t)=\prod_{i=1}^np(\mathbf{c}_i|\mathbf{x}_t),
\end{align*}
Then, we have the summation formulation to rewrite the gradient term in score function
\begin{align}
\nabla_{\mathbf{x}_t}\log p(\mathbf{c}_1,...,\mathbf{c}_n|\mathbf{x}_t)=\sum_{i=1}^n \nabla_{\mathbf{x}_t}\log p(\mathbf{c}_i|\mathbf{x}_t).
\label{eq:cond_independent}
\end{align}
Pioneering work \cite{EGSDE} uses an energy function to approximate $\log p(\mathbf{c}|\mathbf{x}_t)$ and \cite{Freedom, BansalCSSGGG24} empirically proposes to use the summation of multiple energy functions to solve the multi-condition guidance problem, which happens to follow the result in Eq. (\ref{eq:cond_independent}). 

However, the equation in Eq. (\ref{eq:cond_independent}) can not hold owing to the basic probabilistic theory that independence neither implies conditional independence
\begin{align*}
    p(\mathbf{c}_1,...,\mathbf{c}_n)=\prod_{i=1}^np(\mathbf{c}_i)\nRightarrow p(\mathbf{c}_1,...,\mathbf{c}_n|\mathbf{x}_t)=\prod_{i=1}^np(\mathbf{c}_i|\mathbf{x}_t).
\end{align*}
In fact, the noisy data $\mathbf{x}_t$ for $\forall t<n$ contains the information of all conditions involved in the inverse process. For example, if we have already known that $\mathbf{x}_t$ reflects the image of artwork ``Sunflowers" \footnote{Sunflowers is the title of two series of still life paintings by the Dutch painter Vincent van Gogh.} polluted by noise, then knowing that $\mathbf{c}_1$ refers to the object of image is sunflowers tells you a lot about the van Gogh's art style, \textit{i.e.}, $\mathbf{c}_2$. Hence, $p(\mathbf{c}_1|\mathbf{x}_t)$ and $p(\mathbf{c}_2|\mathbf{x}_t)$ are not conditionally independent.

\begin{figure}[t]
    \centering
    \begin{subfigure}[b]{0.235\textwidth}
        \includegraphics[width=\linewidth]{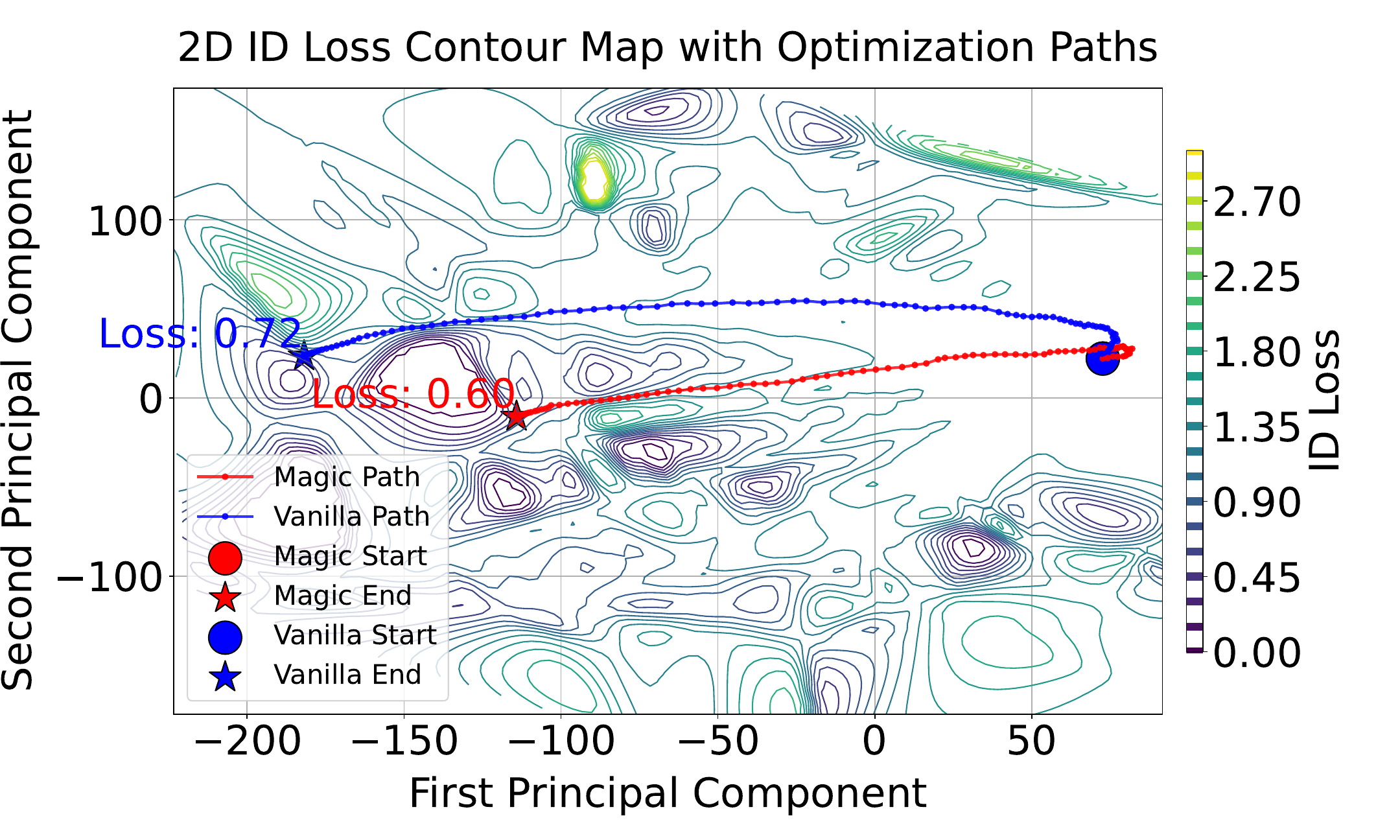}
        \caption{ID Loss}
        \label{fig:id_loss_landscape}
    \end{subfigure}%
    \hfill
    \begin{subfigure}[b]{0.235\textwidth}
        \includegraphics[width=\linewidth]{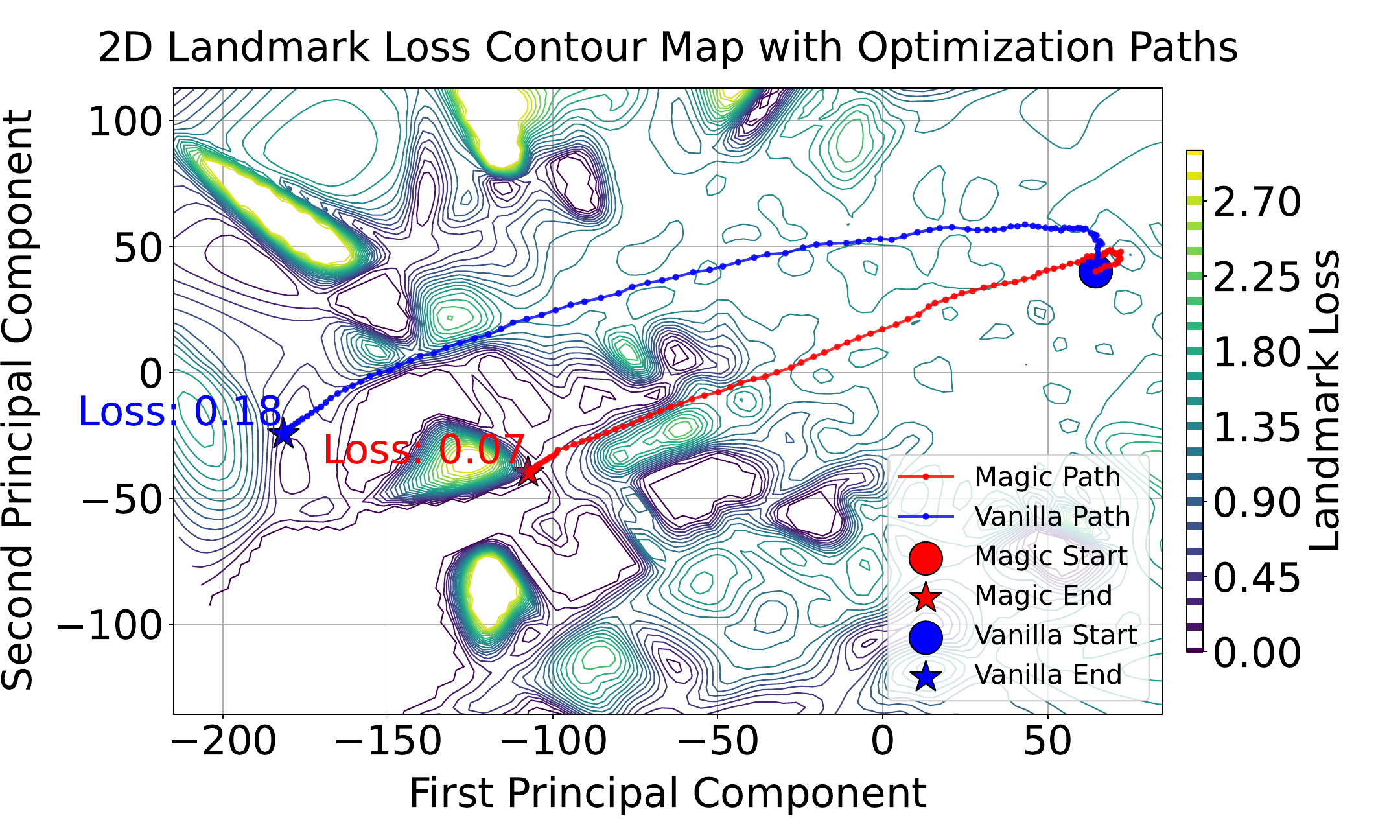}
        \caption{Landmark Loss}
        \label{fig:landmark_loss_landscape}
    \end{subfigure}
    \vspace{-3mm}
    \caption{Illustration of the multi-condition optimization landscape with and without the our strategy (best viewed in color). Our approach $\nabla_{\mathbf{x}_t} \log p(\mathbf{c}_1,\mathbf{c}_2|\mathbf{x}_t)$ navigates the valleys, achieving lower condition losses, whereas the vanilla counterpart $\nabla_{\mathbf{x}_t} \log p(\mathbf{c}_1|\mathbf{x}_t)+\nabla_{\mathbf{x}_t} \log p(\mathbf{c}_2|\mathbf{x}_t)$ requires more steps to find a decreasing direction, resulting in higher loss values.}
    \label{fig:combined_landscape}
\end{figure}

\subsection{Conditional Dependent Creation}
\label{sec:cond_dep}
Since the hypothesis on conditional independence is too strong, we consider the real case of conditional dependence where
\begin{align*}
p(\mathbf{c}_1,...,\mathbf{c}_n|\mathbf{x}_t)&=p(\mathbf{c}_1|\mathbf{x}_t)p(\mathbf{c}_2|\mathbf{c}_1,\mathbf{x}_t)...p(\mathbf{c}_n|\mathbf{c}_1,...,\mathbf{c}_{n-1},\mathbf{x}_t),\\
&=\prod_{i=1}^n p(\mathbf{c}_i|\{\mathbf{c}_{j\in(0,i-1]}\},\mathbf{x}_t).
\end{align*}
To simplify the derivation, we begin with the case with two constraints, i.e., $\{\mathbf{c}_1, \mathbf{c}_2\}$. The controllable term is in the form of 
\begin{align*}
    \nabla_{\mathbf{x}_{t}}\log p(\mathbf{c}_1,\mathbf{c}_2|\mathbf{x}_{t})=\underbrace{\nabla_{\mathbf{x}_{t}}\log p(\mathbf{c}_1|\mathbf{x}_{t})}_{\text{term I}}+\underbrace{\nabla_{\mathbf{x}_{t}}\log p(\mathbf{c}_2|\mathbf{c}_1,\mathbf{x}_{t})}_{\text{term II}},
\end{align*}
where the first term serves as the single condition term as discussed in previous work, and the second term relies on the $\mathbf{c}_1$ guided result, which inspires us first to calculate the intermediate $\hat{\mathbf{x}}_{t,\mathbf{c}_1}$ as 
\begin{align*}
    \hat{\mathbf{x}}_{t,\mathbf{c}_1}=(2-\sqrt{1-\beta_t})\mathbf{x}_t+&\beta_t\big(\mathbf{s}_{\boldsymbol{\theta}^*}(\mathbf{x}_t,t)\\
    &+\nabla_{\mathbf{x}_{t}}\log p(\mathbf{c}_1|\mathbf{x}_{t})\big)+\sqrt{\beta_t}\boldsymbol{\epsilon},
\end{align*}
then use $\hat{\mathbf{x}}_{t,\mathbf{c}_1}$ to derive $\nabla_{\mathbf{x}_{t}}\log p(\mathbf{c}_2|\mathbf{c}_1,\mathbf{x}_{t})\in\mathbb{R}^{1\times HW}$ as 
\begin{align*}
    \nabla_{\mathbf{x}_{t}}\log p(\mathbf{c}_2|\mathbf{c}_1,\mathbf{x}_{t})&= \nabla_{\mathbf{x}_{t}}\hat{\mathbf{x}}_{t,\mathbf{c}_1}\cdot \nabla_{\hat{\mathbf{x}}_{t,\mathbf{c}_1}}\log p(\mathbf{c}_2|\hat{\mathbf{x}}_{t,\mathbf{c}_1}),\\
    &=\nabla_{\mathbf{x}_{t}}\hat{\mathbf{x}}_{t,\mathbf{c}_1}\cdot g_{\hat{\mathbf{x}}_{t,\mathbf{c}_1}},
\end{align*}
where $g_{\hat{\mathbf{x}}_{t,\mathbf{c}_1}}$ denotes $\nabla_{\hat{\mathbf{x}}_{t,\mathbf{c}_1}}\log p(\mathbf{c}_2|\hat{\mathbf{x}}_{t,\mathbf{c}_1})$ for short and
\begin{align*}
\nabla_{\mathbf{x}_{t}}\hat{\mathbf{x}}_{t,\mathbf{c}_1}=(2-\sqrt{1-\beta_t})\mathbf{I}+\beta_t\big(\nabla_{\mathbf{x}_{t}}\mathbf{s}_{\boldsymbol{\theta}^*}(\mathbf{x}_t,t)+\mathbf{H}_{\mathbf{x}_{t}}\big).
\end{align*}
It is easy to calculate the Jacobian matrix of $\nabla_{\mathbf{x}_{t}}\mathbf{s}_{\boldsymbol{\theta}^*}(\mathbf{x}_t,t)$ via standard deep learning libraries such as TensorFlow\cite{tensorflow}/PyTorch\cite{pytorch}. However, the Jacobian and Hessian matrix $\mathbf{H}_{\mathbf{x}_{t}}=\nabla_{\mathbf{x}_{t}}^2\log p(\mathbf{c}_1|\mathbf{x}_{t})$ can be the bottleneck of computation cost and memory footprint \footnote{Taking image with $256\times256$ resolution as an example, the Hessian matrix consumes $512^4\times4$ bytes, i.e., $256$GB, which is impractical.}. Fortunately, by utilizing the product of $\nabla_{\mathbf{x}_{t}}\hat{\mathbf{x}}_{t,\mathbf{c}_1}$ and $g_{\hat{\mathbf{x}}_{t,\mathbf{c}_1}}^T$, we can avoid the direct calculation of the Hessian matrix.

To simplify the derivation, we denote $(2-\sqrt{1-\beta_t})\mathbf{I}+\beta_t\nabla_{\mathbf{x}_{t}}\mathbf{s}_{\boldsymbol{\theta}^*}(\mathbf{x}_t,t)$ as matrix $\mathbf{A}$, then 
\begin{align}
\nabla_{\mathbf{x}_{t}}\log p(\mathbf{c}_2|\mathbf{c}_1,\mathbf{x}_{t})=\mathbf{A}\cdot g_{\hat{\mathbf{x}}_{t,\mathbf{c}_1}}+\beta_t\mathbf{H}_{\mathbf{x}_t}\cdot g_{\hat{\mathbf{x}}_{t,\mathbf{c}_1}}.
\label{gradient_of_c2_wrt_c1_xt}
\end{align}
Similar to $g_{\hat{\mathbf{x}}_{t,\mathbf{c}_1}}$, we further rewrite $\nabla_{\mathbf{x}_{t}}\log p(\mathbf{c}_1|\mathbf{x}_{t})$ as $g_{\mathbf{x}_t}$, then we have
\begin{align}
    \mathbf{H}_{\mathbf{x}_t}\cdot g_{\hat{\mathbf{x}}_{t,\mathbf{c}_1}}=\frac{\partial g_{\mathbf{x}_t}^T}{\partial \mathbf{x}_t}\cdot g_{\hat{\mathbf{x}}_{t,\mathbf{c}_1}}&=\frac{\partial g_{\mathbf{x}_t}^T}{\partial \mathbf{x}_t}\cdot g_{\hat{\mathbf{x}}_{t,\mathbf{c}_1}}+g_{\mathbf{x}_t}^T\cdot\frac{\partial g_{\hat{\mathbf{x}}_{t,\mathbf{c}_1}}}{\partial \mathbf{x}_t} \nonumber \\
    &=\frac{\partial (g_{\mathbf{x}_t}^Tg_{\hat{\mathbf{x}}_{t,\mathbf{c}_1}})}{\partial \mathbf{x}_t}.
    \label{replace_Hg_with_gradient_wrt_scalar}
\end{align}
Here, we make use of the trick that $\frac{\partial g_{\hat{\mathbf{x}}_{t,\mathbf{c}_1}}}{\partial \mathbf{x}_t}=\mathbf{0}$, which convert the product between the Hessian matrix and gradient vector to the gradient of $\mathbf{x}_t$ with respect to a scalar $g_{\mathbf{x}_t}^Tg_{\hat{\mathbf{x}}_{t,\mathbf{c}_1}}$. The computing cost and GPU memory footprint can be much lower than the vanilla formulation. The conclusion also holds for $\nabla_{\mathbf{x}_{t}}\big(\mathbf{s}_{\boldsymbol{\theta}^*}(\mathbf{x}_t,t)\cdot g_{\hat{\mathbf{x}}_{t,\mathbf{c}_1}}\big)$ in $\mathbf{A}\cdot g_{\hat{\mathbf{x}}_{t,\mathbf{c}_1}}$. 

We summarize the pipeline with PC sampler \cite{0011SKKEP21} to solve Equation (\ref{eq_sde}) in Algorithm \ref{alg:pc_ddpm}. Note that our approach only modifies the sampling process to apply the multiple attribute generation leaving out the training procedure, which naturally fits the need of zero-shot learning. Figure \ref{fig:combined_landscape} further illustrates the superiority of Algorithm \ref{alg:pc_ddpm} compared to conditional independent generation. We randomly sample ten thousands results from the space of $\mathbf{x}_t$, then project them to 2D via PCA to construct the loss landscape.

\begin{algorithm}
    \small
       \caption{Conditional Sampling}
       \label{alg:pc_ddpm}
        \begin{algorithmic}[1]
           \State $\bfx_N \sim \mcal{N}(\bfzero, \bfI)$
           \For{$t=N$ {\bfseries to} $1$}
           \State $\nabla_{\mathbf{x}_{t}}\log p(\mathbf{x}_{t}|\mathbf{c}_1)\gets \mathbf{s}_{\boldsymbol{\theta}^*}(\mathbf{x}_{t},t)+\nabla_{\mathbf{x}_{t}}\log p(\mathbf{c}_1|\mathbf{x}_{t})$
        \State{$\hat{\mathbf{x}}_{t,\mathbf{c}_1} \gets (2 - \sqrt{1-\beta_{t}})\bfx_{t} + \beta_{t} \nabla_{\mathbf{x}_{t}}\log p(\mathbf{x}_{t}|\mathbf{c}_1)$}
             \State{$\bfz \sim \mcal{N}(\bfzero, \bfI)$}
             \State{$\hat{\mathbf{x}}_{t,\mathbf{c}_1} \gets \hat{\mathbf{x}}_{t,\mathbf{c}_1} + \sqrt{\beta_{t}} \bfz$}
             \State{$\nabla_{\mathbf{x}_{t}}\log p(\mathbf{c}_2|\mathbf{c}_1,\mathbf{x}_{t}) \gets \text{solve Eq.(}\ref{gradient_of_c2_wrt_c1_xt} \text{) with Eq.(}\ref{replace_Hg_with_gradient_wrt_scalar}\text{)} $}
             \State$\nabla_{\mathbf{x}_{t}}\log p(\mathbf{c}_1,\mathbf{c}_2|\mathbf{x}_{t})\gets\nabla_{\mathbf{x}_{t}}\log p(\mathbf{c}_1|\mathbf{x}_{t})+\nabla_{\mathbf{x}_{t}}\log p(\mathbf{c}_2|\mathbf{c}_1,\mathbf{x}_{t})$
             \State $\mathbf{x}_{t-1}\gets(2 - \sqrt{1-\beta_{t}})\bfx_{t} + \beta_{t}\big(\mathbf{s}_{\boldsymbol{\theta}^*}(\mathbf{x}_{t},t)+\nabla_{\mathbf{x}_{t}}\log p(\mathbf{c}_1,\mathbf{c}_2|\mathbf{x}_{t})\big)$
             \State $\mathbf{x}_{t-1}\gets \mathbf{x}_{t-1} + \sqrt{\beta_{t}} \bfz$
           \EndFor
           \State {\bfseries return} $\bfx_0$
        \end{algorithmic}
\end{algorithm}

\begin{figure*}[t]
    \centering
    \begin{subfigure}[b]{0.3\textwidth}
        \includegraphics[width=\linewidth]{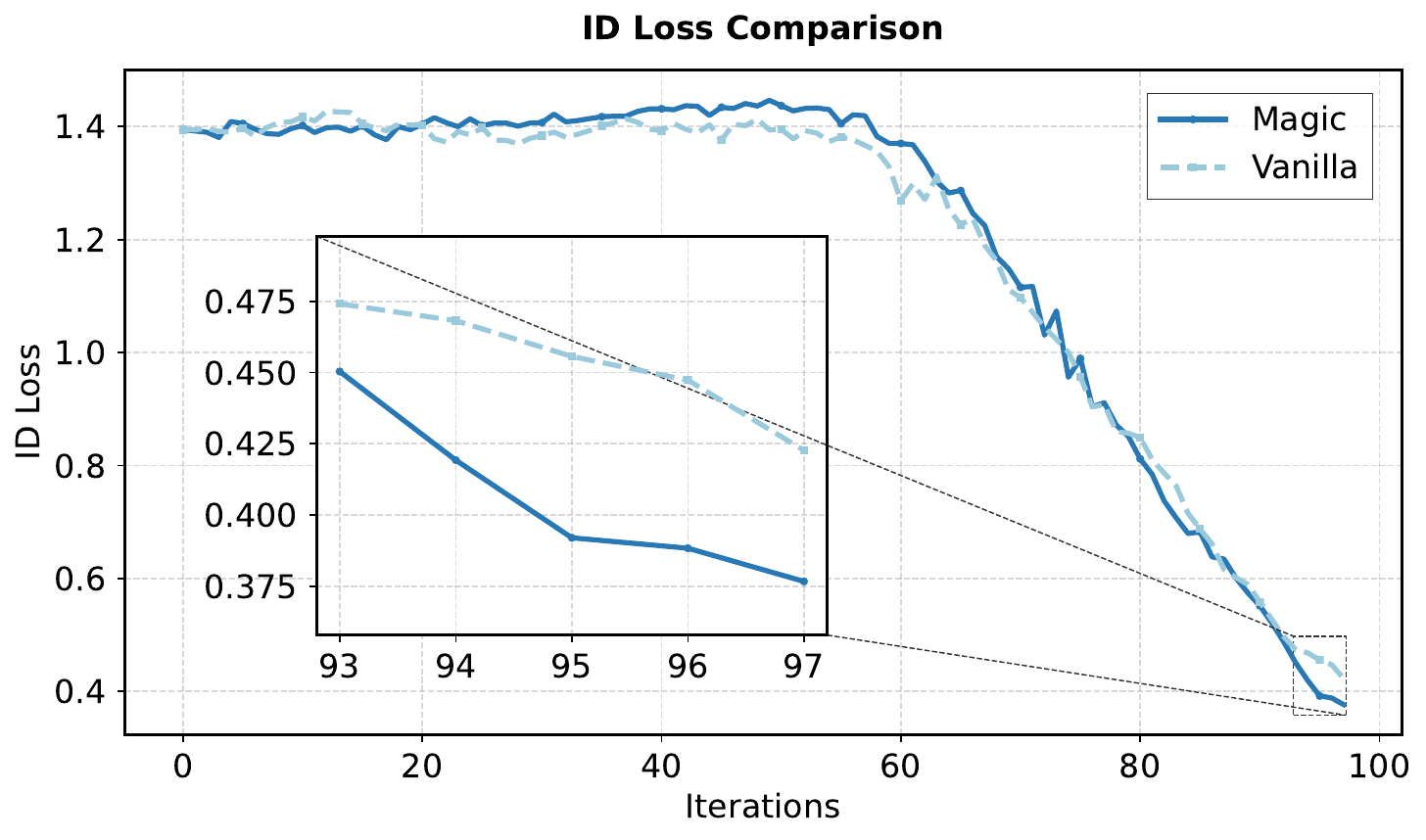}
        \caption{ID Loss Variation Across Iterations}
    \end{subfigure}
    \hfill
    \begin{subfigure}[b]{0.3\textwidth}
        \includegraphics[width=\linewidth]{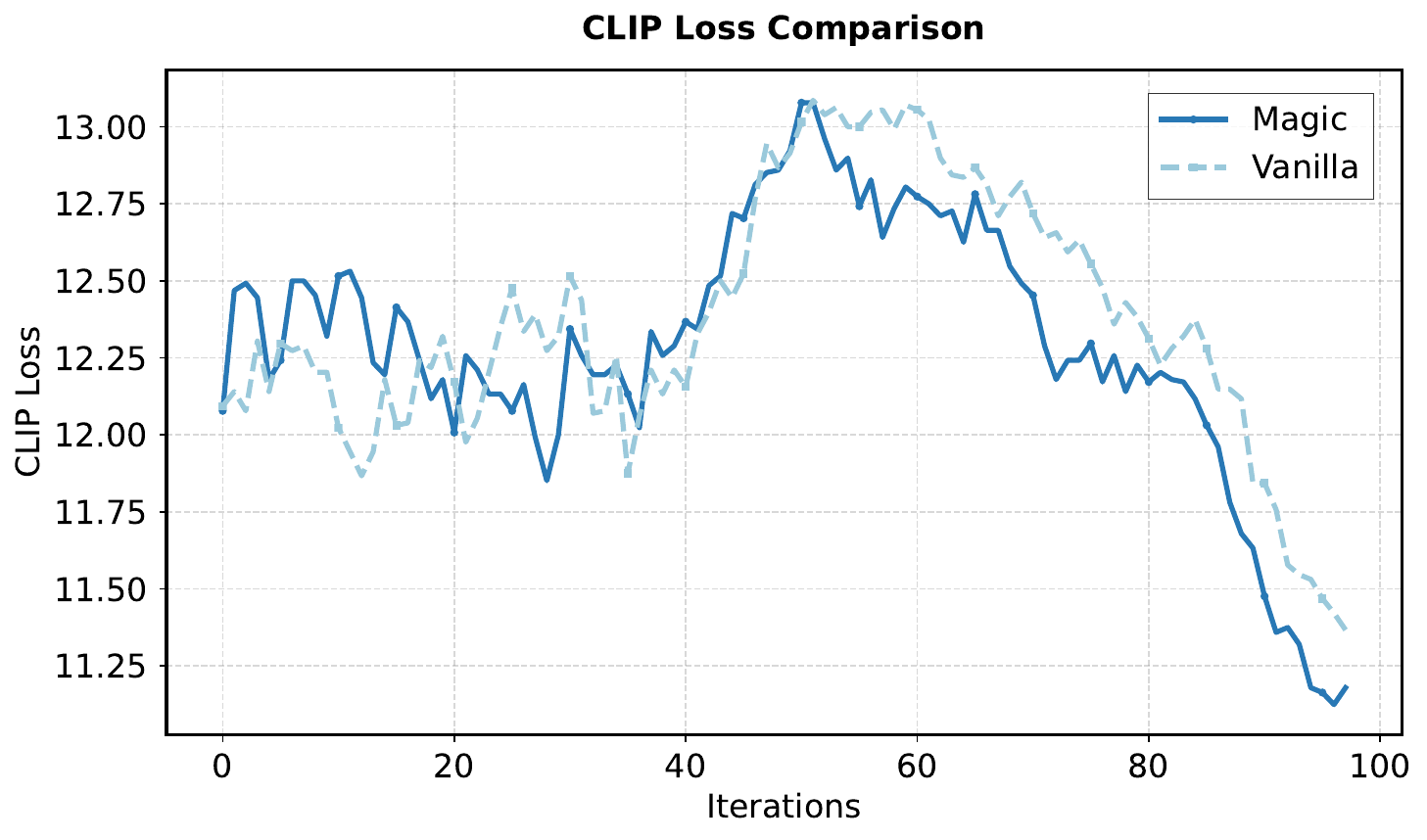}
        \caption{CLIP Loss Variation Across Iterations}
    \end{subfigure}
    \hfill
    \begin{subfigure}[b]{0.3\textwidth}
        \includegraphics[width=\linewidth]{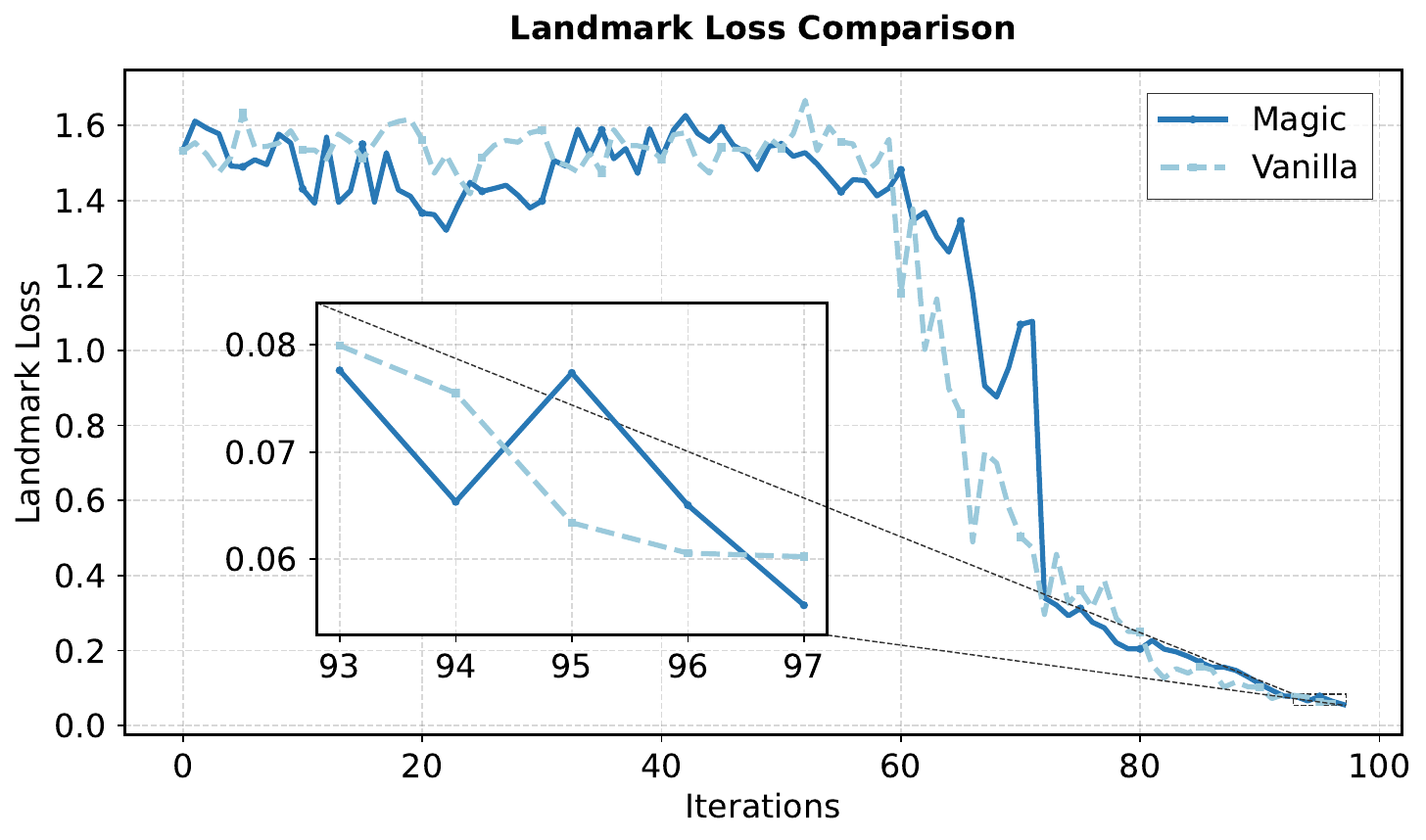}
        \caption{Landmark Loss Variation Across Iterations}
    \end{subfigure}
    \caption{We visualize the loss curves for generating a face image under three guided conditions using $\nabla_{\mathbf{x}_t} \log p(\mathbf{c}_1,\mathbf{c}_2,\mathbf{c}_3|\mathbf{x}_t)$, comparing results with vanilla $\sum_{i}\nabla_{\mathbf{x}_t} \log p(\mathbf{c}_i|\mathbf{x}_t)$. The proposed method achieves relatively lower metric values across all conditions, indicating a better balance in multi-condition control.}
    \label{fig:combined_loss_curve}
\end{figure*}

\subsection{A Multi-task Learning Perspective}
Though Section \ref{sec:cond_dep} alleviates the image customization with two conditions, the complexity for solving $p(\mathbf{c}_i|\mathbf{c}_1,...,\mathbf{c}_{i-1},\mathbf{x}_t)$ still increases with the number of conditions. For example, $\nabla_{\mathbf{x}_{t}}\log p(\mathbf{c}_3|\mathbf{c}_2,\mathbf{c}_1,\mathbf{x}_{t})$ should introduce a third-order derivative, which corresponds to a 3D tensor. Therefore, we propose a simple heuristic method to solve the multi-condition customization.

Given multiple conditions $\{\mathbf{c}_1,...,\mathbf{c}_n\}$, we may go through every $\mathbf{c}_i$ from $\mathbf{c}_1$ to $\mathbf{c}_n$ as the initial condition. Then, follow the derivation in Section \ref{sec:cond_dep}, we obtain $\nabla_{\mathbf{x}_{t}}\log p(\mathbf{c}_j,\mathbf{c}_i|\mathbf{x}_{t})$ for $\forall i,j$. If we regard $\nabla_{\mathbf{x}_{t}}\log p(\mathbf{c}_1,\mathbf{c}_2|\mathbf{x}_{t})$ as the gradient descent step for the loss function $\min -\log p(\mathbf{c}_1,\mathbf{c}_2|\mathbf{x}_{t})$, then the optimal gradient step $g_{\mathbf{x}_t}$ satisfying all losses $\min -\log p(\mathbf{c}_j,\mathbf{c}_i|\mathbf{x}_{t})$ where $\{\forall i,j\in[1,n]\}$ serves as an approximation to $\nabla_{\mathbf{x}_{t}}\log p(\mathbf{c}_1,...,\mathbf{c}_n|\mathbf{x}_{t})$. Formally, the optimal gradient step $g_{\mathbf{x}_t}$ solves the following goal
\begin{align}
    \mathbf{x}_t^*=\arg\min \big\{\mathcal{L}(\mathbf{x}_t)\triangleq \sum_{i}\sum_{j}-\log p(\mathbf{c}_j,\mathbf{c}_i|\mathbf{x}_t) \big\},
\end{align}
which is the standard formulation of Multi-Task Learning (MTL) \cite{VandenhendeGGPD22}. It uses $g_{\mathbf{x}_t}$ to find an optimal $\mathbf{x}_t^*$ that achieves low losses across all tasks.

In this paper, we adopt conflict-averse gradient descent (CAGrad \cite{CAGrad}) to tackle the MTL problem. Algorithm \ref{alg:multi_cond} further illustrates how to approximate $\nabla_{\mathbf{x}_{t}}\log p(\mathbf{c}_1,…,\mathbf{c}_n|\mathbf{x}_{t})$ by CAGrad. Consequently, Figure \ref{fig:combined_loss_curve} demonstrates that our approach performs effectively in multi-condition generation.


\begin{algorithm}[t]
    \small
       \caption{Multi-condition Approximation}
       \label{alg:multi_cond}
        \begin{algorithmic}[1]
           \State $c\in[0,1)$
           \For{$i=1$ {\bfseries to} $n$}
                \For{$j=1$ {\bfseries to} $n$}
                    \State{$\nabla_{\mathbf{x}_{t}}\log p(\mathbf{c}_j|\mathbf{c}_i,\mathbf{x}_{t}) \gets \text{solve Eq.(}\ref{gradient_of_c2_wrt_c1_xt} \text{) with Eq.(}\ref{replace_Hg_with_gradient_wrt_scalar}\text{)} $}
                    \State $\nabla_{\mathbf{x}_{t}}\log p(\mathbf{c}_i,\mathbf{c}_j|\mathbf{x}_{t})\gets\nabla_{\mathbf{x}_{t}}\log p(\mathbf{c}_i|\mathbf{x}_{t})+\nabla_{\mathbf{x}_{t}}\log p(\mathbf{c}_j|\mathbf{c}_i,\mathbf{x}_{t})$
                \EndFor
           \EndFor
           \State $g_0\gets \frac{1}{n(n-1)}\sum_{i}\sum_{j}\nabla_{\mathbf{x}_{t}}\log p(\mathbf{c}_i,\mathbf{c}_j|\mathbf{x}_{t})$
           \State $\phi\gets c^2||g_0||^2$
           \State $g_w\gets \frac{1}{n(n-1)}\sum_{i}\sum_{j}w_{i,j}p(\mathbf{c}_i,\mathbf{c}_j|\mathbf{x}_{t})$
           \State $w^*=\arg\min_{w} g_w^Tg_0+\sqrt{\phi}||g_w||$
           \State {\bfseries return} $g_0+\frac{\sqrt{\phi}}{||g_{w^*}||}g_{w^*}$
        \end{algorithmic}
\end{algorithm}

\subsection{Choice of Condition Classifier $p(\boldsymbol{c}|\boldsymbol{x}_t)$}
The remaining challenge in implementing Algorithms \ref{alg:pc_ddpm} and \ref{alg:multi_cond} lies in determining $p(\mathbf{c}|\mathbf{x}_t)$. A common approach for class-conditional sampling involves employing a pre-trained, time-dependent classifier model for $p(\mathbf{c}|\mathbf{x}_t)$ \cite{DhariwalN21, NicholDRSMMSC22, 0011SKKEP21}, or using a handcrafted energy function for conditional control \cite{LiuPAZCHSRD23, EGSDE}, where $\nabla\mathcal{E}(\mathbf{c},\mathbf{x},t) \propto \nabla \log p(\mathbf{c}|\mathbf{x}_t)$. Recent studies \cite{Freedom, BansalCSSGGG24, ChungKMKY23, LugmayrDRYTG22, ParmarS0LLZ23} leverage the predicted clean image $\mathbf{x}_{0|t}$ to eliminate the dependency on timestamp $t$, using $\nabla\mathcal{E}(\mathbf{c},\mathbf{x}_{0|t})$ as an approximation of $ \nabla \log p(\mathbf{c}|\mathbf{x}_t)$. While our approach is compatible with all these strategies, the latter is particularly advantageous for our multi-attribute customization, as it functions as a plug-and-play module that circumvents the need to explicitly train $p_t(\mathbf{c}_i|x_t)$ with $x_t \sim p_t(\mathbf{x}_t|\mathbf{c}_j)$. Consequently, we focus on using a differentiable guidance function $f$, such as a pre-trained perceptual model for image classification, detection, or segmentation, to represent $\mathcal{E}(\mathbf{c},\mathbf{x}_{0|t})$.

%% file: sec/4_exp.tex
\label{sec:exp}

\section{Experiments}
\subsection{Implementation Details}
We apply our approach to various pre-trained open-source diffusion models to enable zero-shot manipulation of multiple attributes. Following the settings of recent conditional creation methods \cite{ChungKMKY23}, we perform ancestral sampling as proposed in \cite{DDPM} with 100 steps. We focus on multi-modal face generation using an unconditional human face diffusion model \cite{MengHSSWZE22} as the base model and stylized, text-guided image creation based on the stable diffusion model \cite{RombachBLEO22} to ensure fair comparisons with other approaches. Attribute control is achieved by integrating text, segmentation, landmarks, face ID, content, and style information. To address potential challenges with learning rate selection in score-based conditional models, we adopt the same settings reported in \cite{Freedom} to avoid extensive search. As for the formulation of condition classifier, we use the following strategy:

\noindent\textbf{Text:} We employ the pre-trained CLIP model \cite{RadfordKHRGASAM21}, which includes both text and image encoders (with the input size of $224\times224$), to project the generated image and guide-text into a shared feature space. To align these representations, we minimize the cosine similarity between them.

\noindent\textbf{Segmentation:} For segmentation, we utilize the face parsing network \cite{face_parsing_pytorch} to generate segmentation maps for both the input reference image and the denoised image. We then apply Mean Squared Error (MSE) loss to minimize the distance between these segmentation maps.

\noindent\textbf{Landmark:} Using a well-known open-source framework for human face landmark detection \cite{pytorch_face_landmark}, we identify precise landmark positions based on face detection results. As with segmentation, we compute the Euclidean distance between the landmarks on the target and generated images.

\noindent\textbf{Face ID:} To capture discriminative facial features, we leverage ArcFace \cite{DengGYXKZ22}. In line with common practice for facial similarity assessments, we calculate Cosine similarity to determine the match between two faces.

\noindent\textbf{Content:} For content creation, we use the standard v1.4 stable diffusion model \cite{RombachBLEO22} as our base and its output size is $512\times512$. This model generates images guided by a prompt, or if no specific guidance is required, we use a wildcard (``*'') to produce an unconditioned output.

\noindent\textbf{Style:} To ensure stylistic consistency, we draw on the neural style transfer framework \cite{GatysEB15a}. Here, the style similarity is quantified by comparing the Gram matrices produced by the image encoder for both the reference and generated images.

\begin{figure}
\setlength{\abovecaptionskip}{2mm}
\centering
\includegraphics[width= \linewidth]{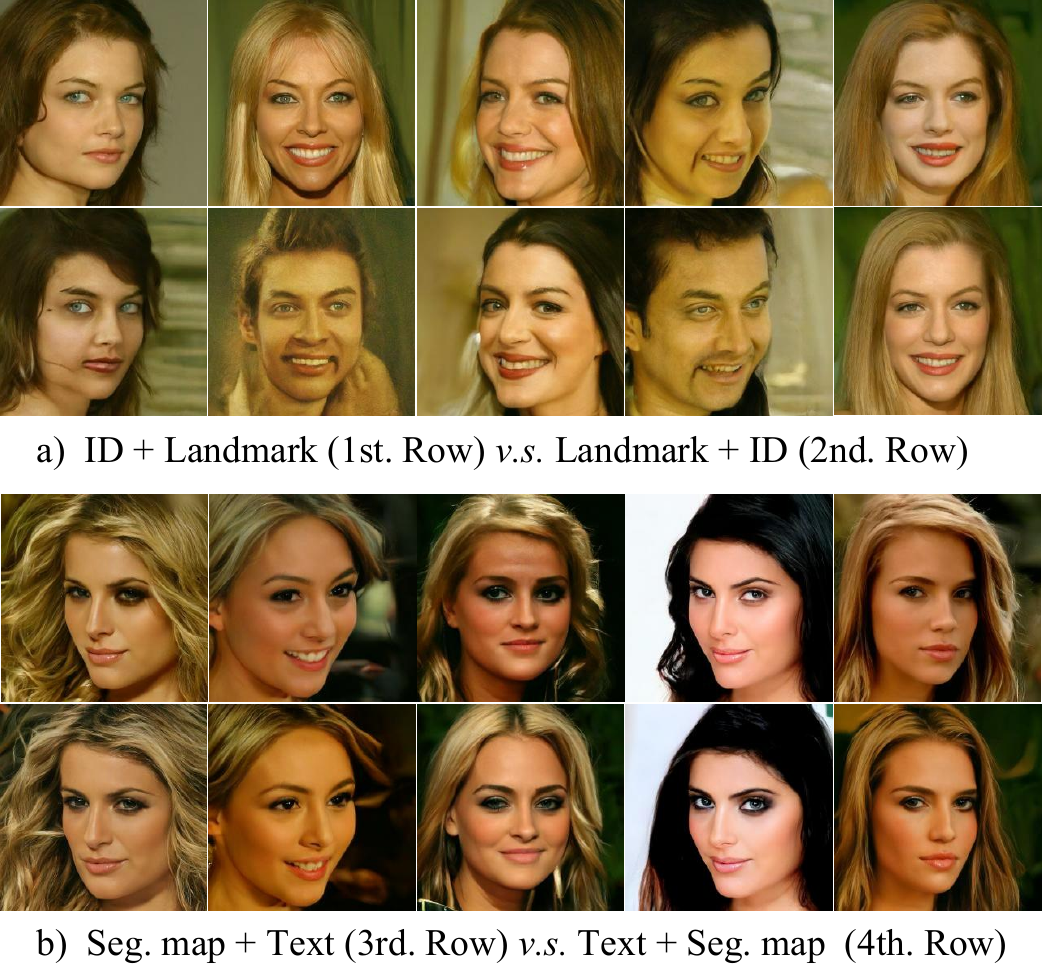}
\caption{Illustration of the effect of different condition sequences. When two conditions modify the same facial features, as shown in a), the sequence becomes important. However, if the conditions are weakly correlated, such as hair color and face parsing, the sequence has minimal impact as shown in b).
}
\label{fig:cond_seq}
\end{figure}

\subsection{Effect of Condition Sequence}
We conduct an ablation study on the impact of using different condition sequences, such as $\{\mathbf{c}_1, \mathbf{c}_2\}$ or $\{\mathbf{c}_2, \mathbf{c}_1\}$, during the diffusion process. Specifically, we examine the differences between applying $\nabla{\mathbf{x}_t} \log p(\mathbf{c}_1|\mathbf{c}_2, \mathbf{x}_t) + \nabla{\mathbf{x}_t} \log p(\mathbf{c}_2|\mathbf{x}_t)$ versus $\nabla{\mathbf{x}_t} \log p(\mathbf{c}_2|\mathbf{c}_1, \mathbf{x}_t) + \nabla{\mathbf{x}_t} \log p(\mathbf{c}_1|\mathbf{x}_t)$ to the denoised intermediate result. As shown in Figure \ref{fig:cond_seq}, the results reveal that if both conditions strongly influence the same facial features, we adhere to the preference of the pre-trained model. For example, when applying face ID and landmark conditions, we find that establishing face ID first for a facial generation model pre-trained on CelebA and then refining landmarks contributes to superior outcomes. If the attributes do not directly conflict, the sequence has minimal effect.

\subsection{Evaluation}
In this section, we present both quantitative and qualitative evaluations. We assess the generated results using qualitative metrics, including Frechet Inception Distance (FID) \cite{HeuselRUNH17} calculated on CelebA-HQ \cite{Lee0W020}, as well as the distance to the given conditions (ID, segmentation, landmarks, and text) using five hundreds randomly selected samples coupled with corresponding reference images, labels and prompts. 

\begin{figure}
\setlength{\abovecaptionskip}{2mm}
\centering
\includegraphics[width= \linewidth]{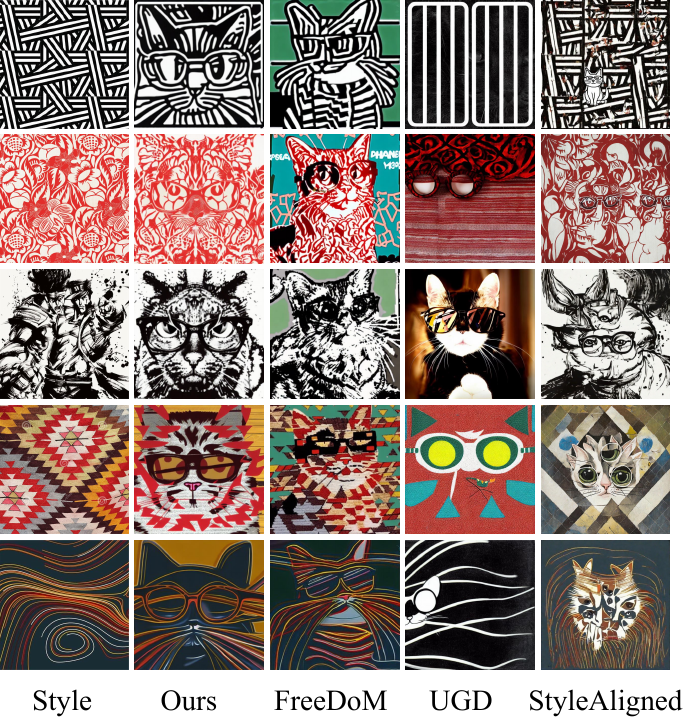}
\caption{Visualization of image creation guided by text and style reference images. All images are generated using the prompt “cat wearing glasses,” with style reference images displayed in the first column.
}
\label{fig:text_style_v1}
\end{figure}

\begin{figure}
\setlength{\abovecaptionskip}{2mm}
\centering
\includegraphics[width= \linewidth]{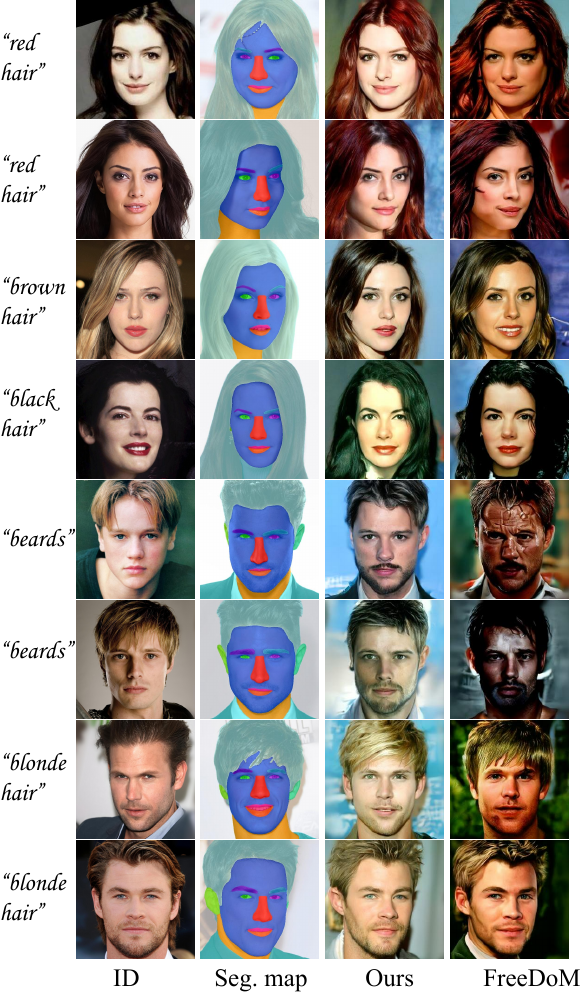}
\caption{Visualization of face synthesis guided by text, face ID and parsing conditions.
}
\label{fig:three_cond}
\end{figure}

\begin{table}
\small
\setlength\tabcolsep{4pt}
\centering
\begin{tabular}{l|c|c|c|c}
\toprule
Method & FID $\downarrow$ & Seg. Dist. $\downarrow$ & ID Dist. $\downarrow$ & Text Dist. $\downarrow$ \\
\midrule
FreeDoM \cite{Freedom} & 136 & 1771 & 0.501 & 0.774 \\
Ours & \textbf{123} & \textbf{1677} & \textbf{0.475} & \textbf{0.769} \\
\bottomrule
\end{tabular}
\vspace{-5pt}
\caption{Quantitative comparison with a condition-independent baseline, evaluating FID, segmentation (Seg.) distance, identity (ID) distance, and text distance. Lower values indicate better performance.}
\label{tab:3cond}
\end{table}

\noindent\textbf{Face Segmentation, ID and Text Control:}
As with dual-condition generation, the independent assumption used in \cite{Freedom} becomes even less realistic in three-condition (text, face parsing and face ID) guided creation. Consequently, our approach (i.e., Algorithm \ref{alg:multi_cond}) outperforms the baseline method, as shown in Figure \ref{fig:three_cond} and Table \ref{tab:3cond}. The text distance appears inconsistent with Figure \ref{fig:combined_loss_curve} because we did not apply condition control in the final steps, resulting in a slightly larger final text distance.

\begin{figure}[t]
\setlength{\abovecaptionskip}{2mm}
\centering
\includegraphics[width= \linewidth]{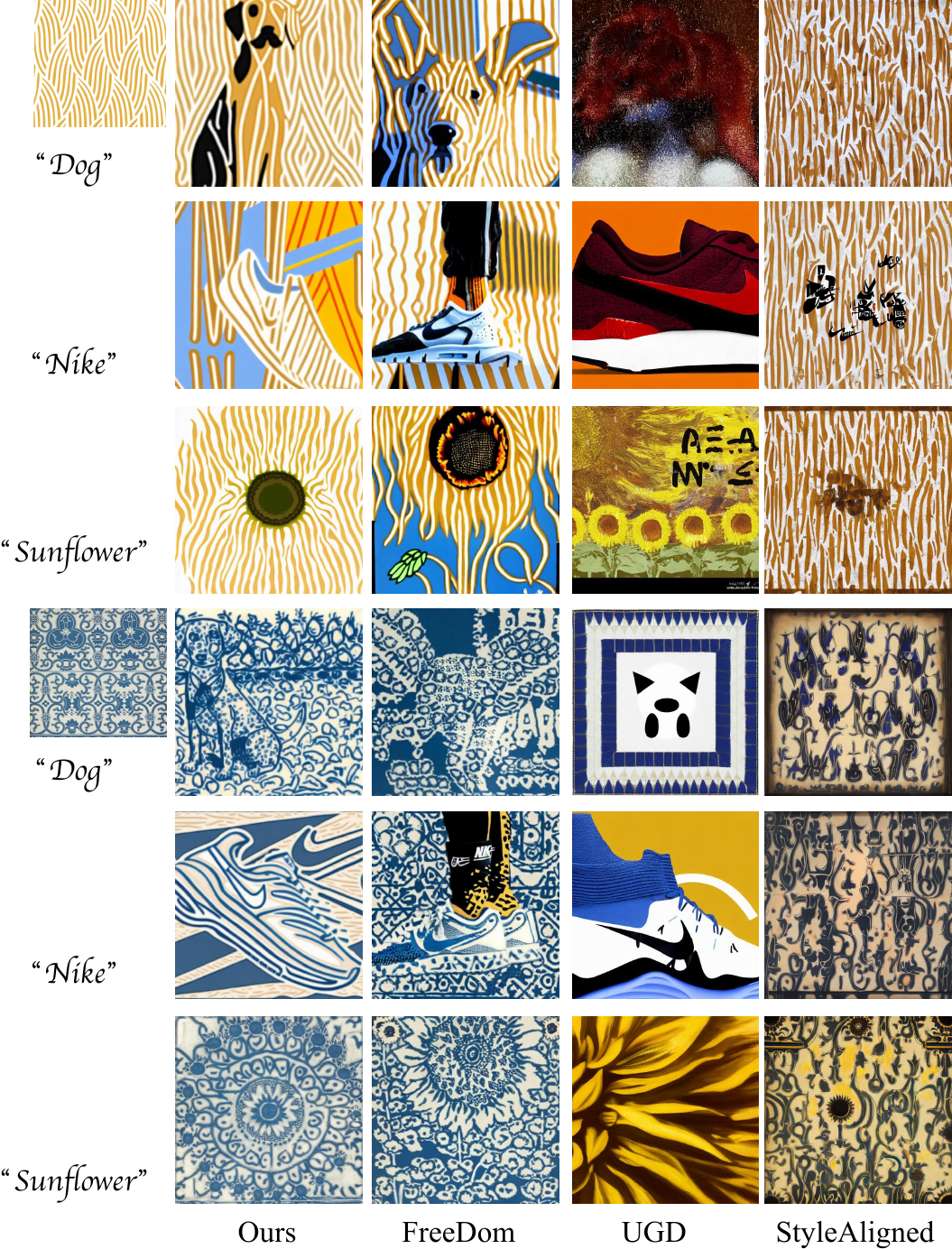}
\caption{Visualization of image creation guided by text and style reference images. We use various prompts and style reference images, shown in the first column, to perform conditional image generation.
}
\label{fig:text_style_v2}
\end{figure}

\noindent\textbf{Text and Style Control:} When coupled with the stable diffusion model, we treat prompt-guided generation as $p(\mathbf{x}_t|\mathbf{c}_1)$ and then apply the style condition $\mathbf{c}_2$ to complete the derivation. As shown in Figures~\ref{fig:text_style_v1} and \ref{fig:text_style_v2}, our approach continues to perform well even under highly abstract style conditions. For comparison, we use two popular methods, UGD \cite{BansalCSSGGG24} and StyleAligned \cite{HertzVFC24}, as baseline methods. Our approach captures the patterns and color schemes of the reference images more effectively, reflecting greater creativity. Quantitative results are shown in Table~\ref{tab:text_style}.

\begin{table}
\small
\setlength\tabcolsep{3pt}
\centering
\begin{tabular}{l|c|c|c}
\toprule
Method & Content Loss $\downarrow$ & Style Loss $\downarrow$ & Text Distance $\downarrow$ \\
\midrule
StyleAligned \cite{HertzVFC24} & - & 11.35 & 0.7475 \\
UGD \cite{BansalCSSGGG24} & - & 18.04 & 0.7682 \\
FreeDoM \cite{Freedom} & 1.93 & 10.21 & 0.7156 \\
Ours & \textbf{1.82} & \textbf{10.14} & \textbf{0.7152}\\
\bottomrule
\end{tabular}
\vspace{-5pt}
\caption{Quantitative comparison with other approaches based on stylized results. We use plain SD-v1.4 results with prompts but without style priors as original content images to calculate content loss. Style loss follows the standard settings in style transfer methods, and text loss is computed using the CLIP model.}
\label{tab:text_style}
\end{table}

\begin{figure}[t]
\setlength{\abovecaptionskip}{2mm}
\centering
\includegraphics[width= \linewidth]{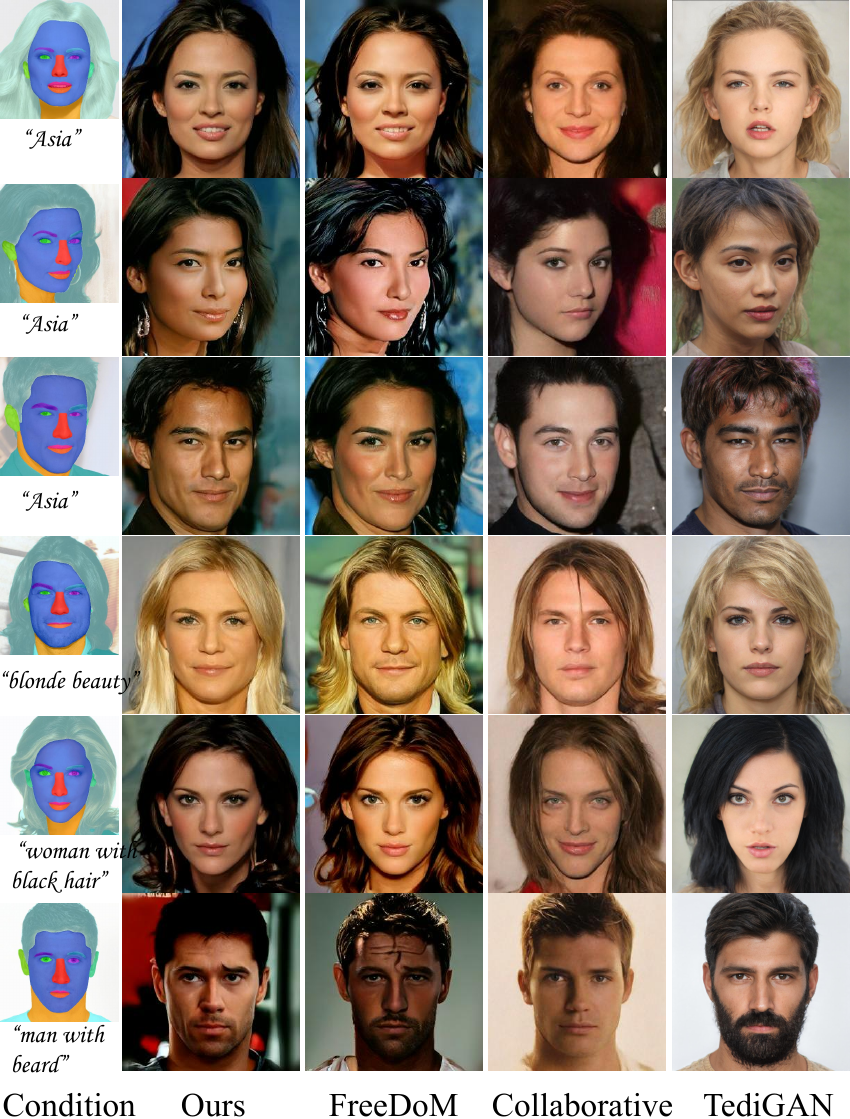}
\caption{Visualization of face synthesis guided by text and face parsing conditions.
}
\label{fig:id_parsing}
\end{figure}

\begin{table}[t]
\small
\setlength\tabcolsep{4pt}
\centering
\begin{tabular}{l|c|c|c}
\toprule
Method & FID $\downarrow$ & Text Distance $\downarrow$ & Seg. Distance $\downarrow$\\
\midrule
TediGAN \cite{XiaYXW21} & \textbf{98} & 0.763 & 2344 \\
Collaborative \cite{HuangC0023} & 122 & 0.758 & 2162 \\
FreeDoM \cite{Freedom} & 136 & 0.764 & 1569\\
Ours & 127 & \textbf{0.752} & \textbf{1502}\\
\bottomrule
\end{tabular}
\vspace{-5pt}
\caption{Quantitative comparison of recent open-source facial conditional generation methods, assessing Fréchet Inception Distance (FID), text embedding distance, and segmentation distance.}
\label{tab:id_parsing}
\end{table}

\noindent\textbf{Face Segmentation and Text Control:}
We compare our method with state-of-the-art approaches, including Collaborative Diffusion \cite{HuangC0023}, TediGAN \cite{XiaYXW21}, and FreeDoM \cite{Freedom}, using segmentation and text guidance as conditions, as shown in Figure \ref{fig:id_parsing} and Table \ref{tab:id_parsing}. Although TediGAN achieves a high FID score, it frequently generates face images that do not match the segmentation mask or deviate from the specified guide text. Collaborative Diffusion performs slightly better but still fails to consistently follow the instructions in certain cases.

\begin{figure}[t]
\setlength{\abovecaptionskip}{2mm}
\centering
\includegraphics[width= \linewidth]{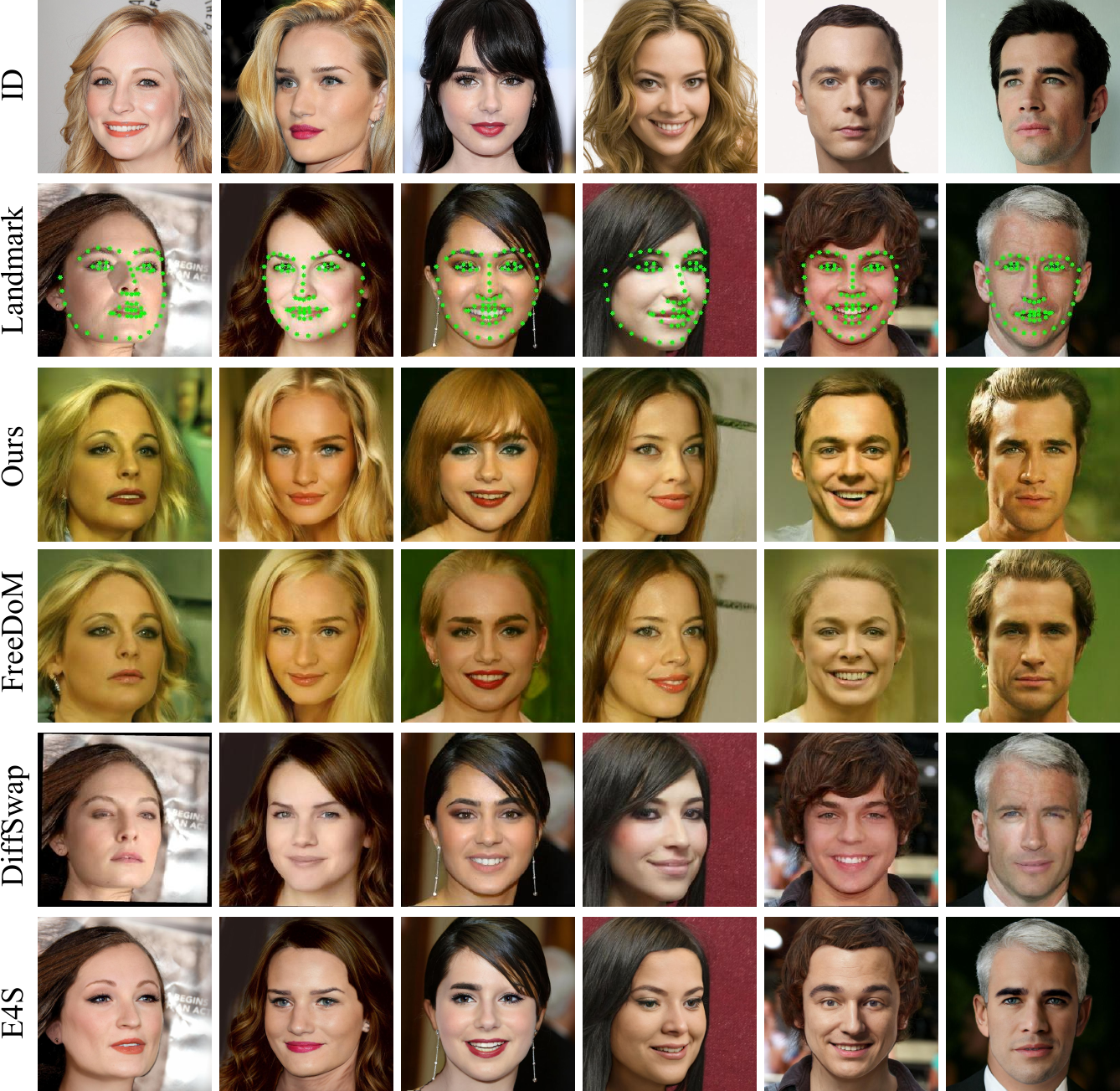}
\caption{Visualization of face synthesis guided by face ID and landmark conditions.
}
\label{fig:id_landmark}
\end{figure}

\noindent\textbf{Face ID and Landmark Control:}
Figure \ref{fig:id_landmark} presents our visual results controlled by Face ID and landmark positions, compared with the conditional independent baseline method, FreeDoM. While FreeDoM can generate face images that align with the input conditions, it often struggles to balance multiple attributes effectively as shown in Table \ref{tab:id_landmark}. 
The SOTA face swapping methods \cite{ZhaoRSL0L23, LiuLZWZWN23} also result in issues such as weakened ID consistency or inaccurate landmark adherence, though, landmark is required during generation. \cite{ZhaoRSL0L23} fails to consider the target face, resulting in a high landmark score, since the original face without ID conditioning should yield a zero landmark distance. Although \cite{LiuLZWZWN23} achieves high-fidelity results, it does so at the cost of partially disregarding both conditions.

\section{Conclusion}
This paper introduced a novel approach for multi-attribute guided generation that explicitly models the conditional dependencies among attributes, enhancing coherence in generated images. Unlike traditional models that assume conditional independence, our method uses a score-based conditional diffusion model to sequentially condition attributes, refining each with contextually-aware gradients. Experimental results demonstrate that Z-Magic effectively overcomes inconsistencies inherent in independent conditioning, achieving high coherence and computational efficiency.

\begin{table}[t]
\small
\setlength\tabcolsep{2.5pt}
\centering
\begin{tabular}{l|c|c|c}
\toprule
Method & FID $\downarrow$ & Landmark Distance $\downarrow$ & ID Distance $\downarrow$ \\
\midrule
DiffSwap~\cite{ZhaoRSL0L23} & 119 & \textbf{0.103} & 1.167 \\
E4S~\cite{LiuLZWZWN23} & \textbf{92} & 0.282 & 0.977 \\
FreeDoM~\cite{Freedom} & 134 & 0.195 & 0.740 \\
Ours & 124 & 0.194 & \textbf{0.549} \\
\bottomrule
\end{tabular}
\vspace{-5pt}
\caption{Quantitative results for the two tasks, ID + Landmark, evaluating FID, landmark distance, and identity (ID) distance. Lower values indicate better performance.}
\label{tab:id_landmark}
\end{table}

%% file: main.bbl
\begin{thebibliography}{51}
\providecommand{\natexlab}[1]{#1}
\providecommand{\url}[1]{\texttt{#1}}
\expandafter\ifx\csname urlstyle\endcsname\relax
  \providecommand{\doi}[1]{doi: #1}\else
  \providecommand{\doi}{doi: \begingroup \urlstyle{rm}\Url}\fi

\bibitem[Abadi et~al.(2016)Abadi, Barham, Chen, Chen, Davis, Dean, Devin, Ghemawat, Irving, Isard, Kudlur, Levenberg, Monga, Moore, Murray, Steiner, Tucker, Vasudevan, Warden, Wicke, Yu, and Zheng]{tensorflow}
Mart{\'{\i}}n Abadi, Paul Barham, Jianmin Chen, Zhifeng Chen, Andy Davis, Jeffrey Dean, Matthieu Devin, Sanjay Ghemawat, Geoffrey Irving, Michael Isard, Manjunath Kudlur, Josh Levenberg, Rajat Monga, Sherry Moore, Derek~Gordon Murray, Benoit Steiner, Paul~A. Tucker, Vijay Vasudevan, Pete Warden, Martin Wicke, Yuan Yu, and Xiaoqiang Zheng.
\newblock Tensorflow: {A} system for large-scale machine learning.
\newblock In \emph{12th {USENIX} Symposium on Operating Systems Design and Implementation, {OSDI} 2016, Savannah, GA, USA, November 2-4, 2016}, pages 265--283. {USENIX} Association, 2016.

\bibitem[Anderson(1982)]{Anderson1982ReversetimeDE}
Brian. D.~O. Anderson.
\newblock Reverse-time diffusion equation models.
\newblock \emph{Stochastic Processes and their Applications}, 12:\penalty0 313--326, 1982.

\bibitem[Avrahami et~al.(2023)Avrahami, Hayes, Gafni, Gupta, Taigman, Parikh, Lischinski, Fried, and Yin]{AvrahamiHGGTPLF23}
Omri Avrahami, Thomas Hayes, Oran Gafni, Sonal Gupta, Yaniv Taigman, Devi Parikh, Dani Lischinski, Ohad Fried, and Xi Yin.
\newblock Spatext: Spatio-textual representation for controllable image generation.
\newblock In \emph{{IEEE/CVF} Conference on Computer Vision and Pattern Recognition, {CVPR} 2023, Vancouver, BC, Canada, June 17-24, 2023}, pages 18370--18380. {IEEE}, 2023.

\bibitem[Bansal et~al.(2024)Bansal, Chu, Schwarzschild, Sengupta, Goldblum, Geiping, and Goldstein]{BansalCSSGGG24}
Arpit Bansal, Hong{-}Min Chu, Avi Schwarzschild, Soumyadip Sengupta, Micah Goldblum, Jonas Geiping, and Tom Goldstein.
\newblock Universal guidance for diffusion models.
\newblock In \emph{The Twelfth International Conference on Learning Representations, {ICLR} 2024, Vienna, Austria, May 7-11, 2024}. OpenReview.net, 2024.

\bibitem[Choi et~al.(2021)Choi, Kim, Jeong, Gwon, and Yoon]{ChoiKJGY21}
Jooyoung Choi, Sungwon Kim, Yonghyun Jeong, Youngjune Gwon, and Sungroh Yoon.
\newblock {ILVR:} conditioning method for denoising diffusion probabilistic models.
\newblock In \emph{2021 {IEEE/CVF} International Conference on Computer Vision, {ICCV} 2021, Montreal, QC, Canada, October 10-17, 2021}, pages 14347--14356. {IEEE}, 2021.

\bibitem[Chung et~al.(2023)Chung, Kim, McCann, Klasky, and Ye]{ChungKMKY23}
Hyungjin Chung, Jeongsol Kim, Michael~Thompson McCann, Marc~Louis Klasky, and Jong~Chul Ye.
\newblock Diffusion posterior sampling for general noisy inverse problems.
\newblock In \emph{The Eleventh International Conference on Learning Representations, {ICLR} 2023, Kigali, Rwanda, May 1-5, 2023}. OpenReview.net, 2023.

\bibitem[cunjian(2023)]{pytorch_face_landmark}
cunjian.
\newblock pytorch\_face\_landmark, 2023.
\newblock Accessed: 2024-11-06.

\bibitem[Deng et~al.(2022{\natexlab{a}})Deng, Guo, Yang, Xue, Kotsia, and Zafeiriou]{DengGYXKZ22}
Jiankang Deng, Jia Guo, Jing Yang, Niannan Xue, Irene Kotsia, and Stefanos Zafeiriou.
\newblock Arcface: Additive angular margin loss for deep face recognition.
\newblock \emph{{IEEE} Trans. Pattern Anal. Mach. Intell.}, 44\penalty0 (10):\penalty0 5962--5979, 2022{\natexlab{a}}.

\bibitem[Deng et~al.(2020)Deng, Tang, Dong, Sun, Huang, and Xu]{deng:2020:arbitrary}
Yingying Deng, Fan Tang, Weiming Dong, Wen Sun, Feiyue Huang, and Changsheng Xu.
\newblock Arbitrary style transfer via multi-adaptation network.
\newblock In \emph{ACM International Conference on Multimedia}, pages 2719--2727, 2020.

\bibitem[Deng et~al.(2021)Deng, Tang, Dong, Huang, Ma, and Xu]{deng:2021:arbitrary}
Yingying Deng, Fan Tang, Weiming Dong, Haibin Huang, Chongyang Ma, and Changsheng Xu.
\newblock Arbitrary video style transfer via multi-channel correlation.
\newblock In \emph{AAAI Conference on Artificial Intelligence (AAAI)}, pages 1210--1217, 2021.

\bibitem[Deng et~al.(2022{\natexlab{b}})Deng, Tang, Dong, Ma, Pan, Wang, and Xu]{Deng:2022:stytr}
Yingying Deng, Fan Tang, Weiming Dong, Chongyang Ma, Xingjia Pan, Lei Wang, and Changsheng Xu.
\newblock Stytr2: Image style transfer with transformers.
\newblock In \emph{Proceedings of the IEEE/CVF Conference on Computer Vision and Pattern Recognition (CVPR)}, pages 11326--11336, 2022{\natexlab{b}}.

\bibitem[Deng et~al.(2024)Deng, He, Tang, and Dong]{DengHTD24}
Yingying Deng, Xiangyu He, Fan Tang, and Weiming Dong.
\newblock Z*: Zero-shot style transfer via attention reweighting.
\newblock In \emph{{IEEE/CVF} Conference on Computer Vision and Pattern Recognition, {CVPR} 2024, Seattle, WA, USA, June 16-22, 2024}, pages 6934--6944. {IEEE}, 2024.

\bibitem[Dhariwal and Nichol(2021)]{DhariwalN21}
Prafulla Dhariwal and Alexander~Quinn Nichol.
\newblock Diffusion models beat gans on image synthesis.
\newblock In \emph{Advances in Neural Information Processing Systems 34: Annual Conference on Neural Information Processing Systems 2021, NeurIPS 2021, December 6-14, 2021, virtual}, pages 8780--8794, 2021.

\bibitem[Gatys et~al.(2015)Gatys, Ecker, and Bethge]{GatysEB15a}
Leon~A. Gatys, Alexander~S. Ecker, and Matthias Bethge.
\newblock A neural algorithm of artistic style.
\newblock \emph{CoRR}, abs/1508.06576, 2015.

\bibitem[Hertz et~al.(2024)Hertz, Voynov, Fruchter, and Cohen{-}Or]{HertzVFC24}
Amir Hertz, Andrey Voynov, Shlomi Fruchter, and Daniel Cohen{-}Or.
\newblock Style aligned image generation via shared attention.
\newblock In \emph{{IEEE/CVF} Conference on Computer Vision and Pattern Recognition, {CVPR} 2024, Seattle, WA, USA, June 16-22, 2024}, pages 4775--4785. {IEEE}, 2024.

\bibitem[Heusel et~al.(2017)Heusel, Ramsauer, Unterthiner, Nessler, and Hochreiter]{HeuselRUNH17}
Martin Heusel, Hubert Ramsauer, Thomas Unterthiner, Bernhard Nessler, and Sepp Hochreiter.
\newblock Gans trained by a two time-scale update rule converge to a local nash equilibrium.
\newblock In \emph{Advances in Neural Information Processing Systems 30: Annual Conference on Neural Information Processing Systems 2017, December 4-9, 2017, Long Beach, CA, {USA}}, pages 6626--6637, 2017.

\bibitem[Ho and Salimans(2022)]{abs-2207-12598}
Jonathan Ho and Tim Salimans.
\newblock Classifier-free diffusion guidance.
\newblock \emph{CoRR}, abs/2207.12598, 2022.

\bibitem[Ho et~al.(2020)Ho, Jain, and Abbeel]{DDPM}
Jonathan Ho, Ajay Jain, and Pieter Abbeel.
\newblock Denoising diffusion probabilistic models.
\newblock In \emph{Advances in Neural Information Processing Systems 33: Annual Conference on Neural Information Processing Systems 2020, NeurIPS 2020, December 6-12, 2020, virtual}, 2020.

\bibitem[Huang et~al.(2023)Huang, Chan, Jiang, and Liu]{HuangC0023}
Ziqi Huang, Kelvin C.~K. Chan, Yuming Jiang, and Ziwei Liu.
\newblock Collaborative diffusion for multi-modal face generation and editing.
\newblock In \emph{{IEEE/CVF} Conference on Computer Vision and Pattern Recognition, {CVPR} 2023, Vancouver, BC, Canada, June 17-24, 2023}, pages 6080--6090. {IEEE}, 2023.

\bibitem[Hyv{\"{a}}rinen(2005)]{Hyvarinen05}
Aapo Hyv{\"{a}}rinen.
\newblock Estimation of non-normalized statistical models by score matching.
\newblock \emph{J. Mach. Learn. Res.}, 6:\penalty0 695--709, 2005.

\bibitem[Karras et~al.(2021)Karras, Laine, and Aila]{KarrasLA21}
Tero Karras, Samuli Laine, and Timo Aila.
\newblock A style-based generator architecture for generative adversarial networks.
\newblock \emph{{IEEE} Trans. Pattern Anal. Mach. Intell.}, 43\penalty0 (12):\penalty0 4217--4228, 2021.

\bibitem[Kim et~al.(2024)Kim, Oh, Do, Kim, and Sohn]{KimODKS24}
Jihyun Kim, Changjae Oh, Hoseok Do, Soohyun Kim, and Kwanghoon Sohn.
\newblock Diffusion-driven {GAN} inversion for multi-modal face image generation.
\newblock In \emph{{IEEE/CVF} Conference on Computer Vision and Pattern Recognition, {CVPR} 2024, Seattle, WA, USA, June 16-22, 2024}, pages 10403--10412. {IEEE}, 2024.

\bibitem[Lee et~al.(2020)Lee, Liu, Wu, and Luo]{Lee0W020}
Cheng{-}Han Lee, Ziwei Liu, Lingyun Wu, and Ping Luo.
\newblock Maskgan: Towards diverse and interactive facial image manipulation.
\newblock In \emph{2020 {IEEE/CVF} Conference on Computer Vision and Pattern Recognition, {CVPR} 2020, Seattle, WA, USA, June 13-19, 2020}, pages 5548--5557. Computer Vision Foundation / {IEEE}, 2020.

\bibitem[Liu et~al.(2021)Liu, Liu, Jin, Stone, and Liu]{CAGrad}
Bo Liu, Xingchao Liu, Xiaojie Jin, Peter Stone, and Qiang Liu.
\newblock Conflict-averse gradient descent for multi-task learning.
\newblock In \emph{Advances in Neural Information Processing Systems 34: Annual Conference on Neural Information Processing Systems 2021, NeurIPS 2021, December 6-14, 2021, virtual}, pages 18878--18890, 2021.

\bibitem[Liu et~al.(2024)Liu, Ma, Zhang, Hu, Fan, Lv, Ding, and Cheng]{LiuM0HFL0C24}
Renshuai Liu, Bowen Ma, Wei Zhang, Zhipeng Hu, Changjie Fan, Tangjie Lv, Yu Ding, and Xuan Cheng.
\newblock Towards a simultaneous and granular identity-expression control in personalized face generation.
\newblock In \emph{{IEEE/CVF} Conference on Computer Vision and Pattern Recognition, {CVPR} 2024, Seattle, WA, USA, June 16-22, 2024}, pages 2114--2123. {IEEE}, 2024.

\bibitem[Liu et~al.(2023{\natexlab{a}})Liu, Park, Azadi, Zhang, Chopikyan, Hu, Shi, Rohrbach, and Darrell]{LiuPAZCHSRD23}
Xihui Liu, Dong~Huk Park, Samaneh Azadi, Gong Zhang, Arman Chopikyan, Yuxiao Hu, Humphrey Shi, Anna Rohrbach, and Trevor Darrell.
\newblock More control for free! image synthesis with semantic diffusion guidance.
\newblock In \emph{{IEEE/CVF} Winter Conference on Applications of Computer Vision, {WACV} 2023, Waikoloa, HI, USA, January 2-7, 2023}, pages 289--299. {IEEE}, 2023{\natexlab{a}}.

\bibitem[Liu et~al.(2023{\natexlab{b}})Liu, Li, Zhang, Wang, Zhang, Wang, and Nie]{LiuLZWZWN23}
Zhian Liu, Maomao Li, Yong Zhang, Cairong Wang, Qi Zhang, Jue Wang, and Yongwei Nie.
\newblock Fine-grained face swapping via regional {GAN} inversion.
\newblock In \emph{{IEEE/CVF} Conference on Computer Vision and Pattern Recognition, {CVPR} 2023, Vancouver, BC, Canada, June 17-24, 2023}, pages 8578--8587. {IEEE}, 2023{\natexlab{b}}.

\bibitem[Lugmayr et~al.(2022)Lugmayr, Danelljan, Romero, Yu, Timofte, and Gool]{LugmayrDRYTG22}
Andreas Lugmayr, Martin Danelljan, Andr{\'{e}}s Romero, Fisher Yu, Radu Timofte, and Luc~Van Gool.
\newblock Repaint: Inpainting using denoising diffusion probabilistic models.
\newblock In \emph{{IEEE/CVF} Conference on Computer Vision and Pattern Recognition, {CVPR} 2022, New Orleans, LA, USA, June 18-24, 2022}, pages 11451--11461. {IEEE}, 2022.

\bibitem[Meng et~al.(2022)Meng, He, Song, Song, Wu, Zhu, and Ermon]{MengHSSWZE22}
Chenlin Meng, Yutong He, Yang Song, Jiaming Song, Jiajun Wu, Jun{-}Yan Zhu, and Stefano Ermon.
\newblock Sdedit: Guided image synthesis and editing with stochastic differential equations.
\newblock In \emph{The Tenth International Conference on Learning Representations, {ICLR} 2022, Virtual Event, April 25-29, 2022}. OpenReview.net, 2022.

\bibitem[Mou et~al.(2024)Mou, Wang, Xie, Wu, Zhang, Qi, and Shan]{MouWXW0QS24}
Chong Mou, Xintao Wang, Liangbin Xie, Yanze Wu, Jian Zhang, Zhongang Qi, and Ying Shan.
\newblock T2i-adapter: Learning adapters to dig out more controllable ability for text-to-image diffusion models.
\newblock In \emph{Thirty-Eighth {AAAI} Conference on Artificial Intelligence, {AAAI} 2024, Thirty-Sixth Conference on Innovative Applications of Artificial Intelligence, {IAAI} 2024, Fourteenth Symposium on Educational Advances in Artificial Intelligence, {EAAI} 2014, February 20-27, 2024, Vancouver, Canada}, pages 4296--4304. {AAAI} Press, 2024.

\bibitem[Nichol et~al.(2022)Nichol, Dhariwal, Ramesh, Shyam, Mishkin, McGrew, Sutskever, and Chen]{NicholDRSMMSC22}
Alexander~Quinn Nichol, Prafulla Dhariwal, Aditya Ramesh, Pranav Shyam, Pamela Mishkin, Bob McGrew, Ilya Sutskever, and Mark Chen.
\newblock {GLIDE:} towards photorealistic image generation and editing with text-guided diffusion models.
\newblock In \emph{International Conference on Machine Learning, {ICML} 2022, 17-23 July 2022, Baltimore, Maryland, {USA}}, pages 16784--16804. {PMLR}, 2022.

\bibitem[Parmar et~al.(2023)Parmar, Singh, Zhang, Li, Lu, and Zhu]{ParmarS0LLZ23}
Gaurav Parmar, Krishna~Kumar Singh, Richard Zhang, Yijun Li, Jingwan Lu, and Jun{-}Yan Zhu.
\newblock Zero-shot image-to-image translation.
\newblock In \emph{{ACM} {SIGGRAPH} 2023 Conference Proceedings, {SIGGRAPH} 2023, Los Angeles, CA, USA, August 6-10, 2023}, pages 11:1--11:11. {ACM}, 2023.

\bibitem[Paszke et~al.(2019)Paszke, Gross, Massa, Lerer, Bradbury, Chanan, Killeen, Lin, Gimelshein, Antiga, Desmaison, K{\"{o}}pf, Yang, DeVito, Raison, Tejani, Chilamkurthy, Steiner, Fang, Bai, and Chintala]{pytorch}
Adam Paszke, Sam Gross, Francisco Massa, Adam Lerer, James Bradbury, Gregory Chanan, Trevor Killeen, Zeming Lin, Natalia Gimelshein, Luca Antiga, Alban Desmaison, Andreas K{\"{o}}pf, Edward~Z. Yang, Zachary DeVito, Martin Raison, Alykhan Tejani, Sasank Chilamkurthy, Benoit Steiner, Lu Fang, Junjie Bai, and Soumith Chintala.
\newblock Pytorch: An imperative style, high-performance deep learning library.
\newblock In \emph{Advances in Neural Information Processing Systems 32: Annual Conference on Neural Information Processing Systems 2019, NeurIPS 2019, December 8-14, 2019, Vancouver, BC, Canada}, pages 8024--8035, 2019.

\bibitem[Podell et~al.(2024)Podell, English, Lacey, Blattmann, Dockhorn, M{\"{u}}ller, Penna, and Rombach]{PodellELBDMPR24}
Dustin Podell, Zion English, Kyle Lacey, Andreas Blattmann, Tim Dockhorn, Jonas M{\"{u}}ller, Joe Penna, and Robin Rombach.
\newblock {SDXL:} improving latent diffusion models for high-resolution image synthesis.
\newblock In \emph{The Twelfth International Conference on Learning Representations, {ICLR} 2024, Vienna, Austria, May 7-11, 2024}. OpenReview.net, 2024.

\bibitem[Radford et~al.(2021)Radford, Kim, Hallacy, Ramesh, Goh, Agarwal, Sastry, Askell, Mishkin, Clark, Krueger, and Sutskever]{RadfordKHRGASAM21}
Alec Radford, Jong~Wook Kim, Chris Hallacy, Aditya Ramesh, Gabriel Goh, Sandhini Agarwal, Girish Sastry, Amanda Askell, Pamela Mishkin, Jack Clark, Gretchen Krueger, and Ilya Sutskever.
\newblock Learning transferable visual models from natural language supervision.
\newblock In \emph{Proceedings of the 38th International Conference on Machine Learning, {ICML} 2021, 18-24 July 2021, Virtual Event}, pages 8748--8763. {PMLR}, 2021.

\bibitem[Ramesh et~al.(2021)Ramesh, Pavlov, Goh, Gray, Voss, Radford, Chen, and Sutskever]{RameshPGGVRCS21}
Aditya Ramesh, Mikhail Pavlov, Gabriel Goh, Scott Gray, Chelsea Voss, Alec Radford, Mark Chen, and Ilya Sutskever.
\newblock Zero-shot text-to-image generation.
\newblock In \emph{Proceedings of the 38th International Conference on Machine Learning, {ICML} 2021, 18-24 July 2021, Virtual Event}, pages 8821--8831. {PMLR}, 2021.

\bibitem[Ramesh et~al.(2022)Ramesh, Dhariwal, Nichol, Chu, and Chen]{abs-2204-06125}
Aditya Ramesh, Prafulla Dhariwal, Alex Nichol, Casey Chu, and Mark Chen.
\newblock Hierarchical text-conditional image generation with {CLIP} latents.
\newblock \emph{CoRR}, abs/2204.06125, 2022.

\bibitem[Rombach et~al.(2022)Rombach, Blattmann, Lorenz, Esser, and Ommer]{RombachBLEO22}
Robin Rombach, Andreas Blattmann, Dominik Lorenz, Patrick Esser, and Bj{\"{o}}rn Ommer.
\newblock High-resolution image synthesis with latent diffusion models.
\newblock In \emph{{IEEE/CVF} Conference on Computer Vision and Pattern Recognition, {CVPR} 2022, New Orleans, LA, USA, June 18-24, 2022}, pages 10674--10685. {IEEE}, 2022.

\bibitem[Shewchuk(1998)]{Shewchuk1998}
Jonathan~Richard Shewchuk.
\newblock High dimensions; random projection; the pseudoinverse, 1998.
\newblock \url{https://people.eecs.berkeley.edu/~jrs/189/lec/22.pdf}.

\bibitem[Song et~al.(2019)Song, Garg, Shi, and Ermon]{SongGSE19}
Yang Song, Sahaj Garg, Jiaxin Shi, and Stefano Ermon.
\newblock Sliced score matching: {A} scalable approach to density and score estimation.
\newblock In \emph{Proceedings of the Thirty-Fifth Conference on Uncertainty in Artificial Intelligence, {UAI} 2019, Tel Aviv, Israel, July 22-25, 2019}, pages 574--584. {AUAI} Press, 2019.

\bibitem[Song et~al.(2021{\natexlab{a}})Song, Durkan, Murray, and Ermon]{SongDME21}
Yang Song, Conor Durkan, Iain Murray, and Stefano Ermon.
\newblock Maximum likelihood training of score-based diffusion models.
\newblock In \emph{Advances in Neural Information Processing Systems 34: Annual Conference on Neural Information Processing Systems 2021, NeurIPS 2021, December 6-14, 2021, virtual}, pages 1415--1428, 2021{\natexlab{a}}.

\bibitem[Song et~al.(2021{\natexlab{b}})Song, Sohl{-}Dickstein, Kingma, Kumar, Ermon, and Poole]{0011SKKEP21}
Yang Song, Jascha Sohl{-}Dickstein, Diederik~P. Kingma, Abhishek Kumar, Stefano Ermon, and Ben Poole.
\newblock Score-based generative modeling through stochastic differential equations.
\newblock In \emph{9th International Conference on Learning Representations, {ICLR} 2021, Virtual Event, Austria, May 3-7, 2021}. OpenReview.net, 2021{\natexlab{b}}.

\bibitem[Vandenhende et~al.(2022)Vandenhende, Georgoulis, Gansbeke, Proesmans, Dai, and Gool]{VandenhendeGGPD22}
Simon Vandenhende, Stamatios Georgoulis, Wouter~Van Gansbeke, Marc Proesmans, Dengxin Dai, and Luc~Van Gool.
\newblock Multi-task learning for dense prediction tasks: {A} survey.
\newblock \emph{{IEEE} Trans. Pattern Anal. Mach. Intell.}, 44\penalty0 (7):\penalty0 3614--3633, 2022.

\bibitem[Wang et~al.(2024)Wang, Darrell, Rambhatla, Girdhar, and Misra]{0007DRGM24}
Xudong Wang, Trevor Darrell, Sai~Saketh Rambhatla, Rohit Girdhar, and Ishan Misra.
\newblock Instancediffusion: Instance-level control for image generation.
\newblock In \emph{{IEEE/CVF} Conference on Computer Vision and Pattern Recognition, {CVPR} 2024, Seattle, WA, USA, June 16-22, 2024}, pages 6232--6242. {IEEE}, 2024.

\bibitem[Wei et~al.(2022)Wei, Chen, Zhou, Liao, Tan, Yuan, Zhang, and Yu]{Wei0Z0TY0Y22}
Tianyi Wei, Dongdong Chen, Wenbo Zhou, Jing Liao, Zhentao Tan, Lu Yuan, Weiming Zhang, and Nenghai Yu.
\newblock Hairclip: Design your hair by text and reference image.
\newblock In \emph{{IEEE/CVF} Conference on Computer Vision and Pattern Recognition, {CVPR} 2022, New Orleans, LA, USA, June 18-24, 2022}, pages 18051--18060. {IEEE}, 2022.

\bibitem[Xia et~al.(2021)Xia, Yang, Xue, and Wu]{XiaYXW21}
Weihao Xia, Yujiu Yang, Jing{-}Hao Xue, and Baoyuan Wu.
\newblock Tedigan: Text-guided diverse face image generation and manipulation.
\newblock In \emph{{IEEE} Conference on Computer Vision and Pattern Recognition, {CVPR} 2021, virtual, June 19-25, 2021}, pages 2256--2265. Computer Vision Foundation / {IEEE}, 2021.

\bibitem[Yu et~al.(2023)Yu, Wang, Zhao, Ghanem, and Zhang]{Freedom}
Jiwen Yu, Yinhuai Wang, Chen Zhao, Bernard Ghanem, and Jian Zhang.
\newblock Freedom: Training-free energy-guided conditional diffusion model.
\newblock In \emph{{IEEE/CVF} International Conference on Computer Vision, {ICCV} 2023, Paris, France, October 1-6, 2023}, pages 23117--23127. {IEEE}, 2023.

\bibitem[Zhang et~al.(2023)Zhang, Rao, and Agrawala]{ZhangRA23}
Lvmin Zhang, Anyi Rao, and Maneesh Agrawala.
\newblock Adding conditional control to text-to-image diffusion models.
\newblock In \emph{{IEEE/CVF} International Conference on Computer Vision, {ICCV} 2023, Paris, France, October 1-6, 2023}, pages 3813--3824. {IEEE}, 2023.

\bibitem[Zhao et~al.(2022)Zhao, Bao, Li, and Zhu]{EGSDE}
Min Zhao, Fan Bao, Chongxuan Li, and Jun Zhu.
\newblock {EGSDE:} unpaired image-to-image translation via energy-guided stochastic differential equations.
\newblock In \emph{Advances in Neural Information Processing Systems 35: Annual Conference on Neural Information Processing Systems 2022, NeurIPS 2022, New Orleans, LA, USA, November 28 - December 9, 2022}, 2022.

\bibitem[Zhao et~al.(2023)Zhao, Rao, Shi, Liu, Zhou, and Lu]{ZhaoRSL0L23}
Wenliang Zhao, Yongming Rao, Weikang Shi, Zuyan Liu, Jie Zhou, and Jiwen Lu.
\newblock Diffswap: High-fidelity and controllable face swapping via 3d-aware masked diffusion.
\newblock In \emph{{IEEE/CVF} Conference on Computer Vision and Pattern Recognition, {CVPR} 2023, Vancouver, BC, Canada, June 17-24, 2023}, pages 8568--8577. {IEEE}, 2023.

\bibitem[zllrunning(2023)]{face_parsing_pytorch}
zllrunning.
\newblock face-parsing.pytorch, 2023.
\newblock Accessed: 2024-11-06.

\end{thebibliography}
